# Generative retrieval-augmented ontologic graph and multi-agent strategies for interpretive large language model-based materials design

Markus J. Buehler[1,2,3,4*]

[1] Laboratory for Atomistic and Molecular Mechanics (LAMM), Massachusetts Institute of Technology, 77 Massachusetts Ave., Cambridge, MA 02139, USA

[2] Department of Civil and Environmental Engineering, Massachusetts Institute of Technology, 77 Massachusetts Ave., Cambridge, MA 02139, USA

[3] Department of Mechanical Engineering, Massachusetts Institute of Technology, 77 Massachusetts Ave., Cambridge, MA 02139, USA

[4] Center for Computational Science and Engineering, Schwarzman College of Computing, Massachusetts Institute of Technology, 77 Massachusetts Ave., Cambridge, MA 02139, USA

[*]E-mail: mbuehler@mit.edu

**Abstract:** Transformer neural networks show promising capabilities, in particular for uses in materials analysis, design and manufacturing, including their capacity to work effectively with both human language, symbols, code, and numerical data. Here we explore the use of large language models (LLMs) as a tool that can support engineering analysis of materials, applied to retrieving key information about subject areas, developing research hypotheses, discovery of mechanistic relationships across disparate areas of knowledge, and writing and executing simulation codes for active knowledge generation based on physical ground truths. When used as sets of AI agents with specific features, capabilities, and instructions, LLMs can provide powerful problem solution strategies for applications in analysis and design problems. Our experiments focus on using a fine-tuned model, MechGPT, developed based on training data in the mechanics of materials domain. We first affirm how finetuning endows LLMs with reasonable understanding of domain knowledge. However, when queried outside the context of learned matter, LLMs can have difficulty to recall correct information. We show how this can be addressed using retrieval-augmented Ontological Knowledge Graph strategies that discern how the model understands what concepts are important and how they are related. Illustrated for a use case of relating distinct areas of knowledge – here, music and proteins – such strategies can also provide an interpretable graph structure with rich information at the node, edge and subgraph level. We discuss nonlinear sampling strategies and agent-based modeling applied to complex question answering, code generation and execution in the context of automated force field development from actively learned Density Functional Theory (DFT) modeling, and data analysis.

## 1. Introduction

Large language models (LLM) based on the general framework of decoder-only autoregressive transformer models [1–6] provide powerful tools for scientific exploration. The specific class of Generative Pretrained Transformer (GPT) models have received significant attention across many fields of inquiry, suggesting remarkable possibilities especially for generative forward and inverse tasks [5,7–15]. However, our understanding of their behavior is still in its infancy, and issues remain with respect to fact recall and potential hallucination, which requires careful validation of their predictions and a thorough exploration of societal implications [16–23]. There have been several recent developments that propose the use of LLM-type models [24–26], and more generally attention-based transformer architectures, to capture the behavior of physical systems including materials [27–30]. Other studies have suggested broad applicability of transformer models in use cases ranging from protein folding [31], protein property prediction [32–36], mechanical field predictions [37,38], as optimizers [39], materials design [14,28,29,40,41], and as educational tools [42–44], among many others. The availability of large models is facilitated by recent developments such as the GPT-4 model [45], and also releases of open-source LLMs such as Llama-2 [46] or Falcon (where a recent release featured a 180 billion parameter model that likely rivals closed-source models such as GPT-3.5) [47,48]. Here we discuss strategies for how LLMs can be improved to provide more accurate responses especially in the context of materials analysis and design, and what mechanisms we can use to elicit more nuanced, detailed and relevant outcomes including interpretability. The strengths of such models reside in part due to their intrinsic ability to learn structurally, as transformer models are ultimately graph-forming models; and when combined into sets of autonomously interacting agents they can form deep networks of interacting systems to solve complex problems that can not only retrieve baked-in information from training, but retrieve knowledge from sources, internet searches, and also write code to run first-principles atomistic simulations by connecting disparate simulation engines [49]. The intrinsic graph-forming capacity can thereby be augmented by amending such mechanistic strategies at higher levels in the materiomic space of structure-property relationships by discovering and using ontological principles akin to categorization [50–58]. This then provides a fertile environment to solve a variety of materials analysis problems including design.

We build this work on a recently developed fine-tuned MechGPT model [30,59] that was trained on a dataset of domain knowledge in the area of mechanics of materials, but was extended here by including a larger training set to also feature Wikipedia articles related to mechanics content as well as textbooks on materials failure [60] (details see **Materials and Methods**). The purpose of this study is to explore the behavior of the model in its trained form, and enhancing generative capabilities by using complex prompting to elicit more accurate responses, more context and featuring higher levels of details. We also discuss the role of nonlinear prompting strategies where LLMs are used to generate concepts, then judge these concepts against a task, and incorporate these various insights to formulate a final response. We explore specifically Retrieval Augmented Generation (RAG) strategies, both using embedding-based indexing of relevant context and via the use of Ontological Knowledge Graphs that offers a deeper mechanistic delineation of knowledge akin to ontology logs [38,54,56–58,61]. In such computational approaches we augment the prompt with relevant information so that the model has access to expanded context that can include details, measurements or new data that thereby greatly expand the capabilities of a LLM during generation. We demonstrate these tools against a series of general mechanics/failure knowledge as relevant in a materials design context to show how a fine-tuned LLM combined with Ontological Knowledge Graphs strategies can be a particularly powerful combination to expand the use cases of LLMs in scientific applications that features both accuracy and interpretability.

The plan for this paper is to first introduce the general method, show some of the limitations of conventional LLMs, then move on to the presentation of key ideas and results, agent-based modeling, materials-design applications including automatic force-field fitting, followed by a discussion. We conclude with an outlook to future research especially in the context of materials design.

## 2. Results and Discussion

LLMs are powerful tools to answer questions, develop hypothesis, summarize and contrast ideas and content, and process complex information, tasks and write code. In the following sections we go through a series of computational experiments to probe the performance of LLMs in the various domains, starting with simple retrieval tasks, moving on to graph-based strategies, and ultimately using multi-agent AI systems where several models interact via a group chat to solve problems as a team.

### 2.1 General query strategies

We explore first parameter-based knowledge in the fine-tuned LLM, with results to a series of questions shown in **Table 1**. As can be seen the model does a reasonable job overall and accurately answers complex domain-specific question. This is in general agreement with results published in earlier papers [28,59]. Next, we run the same experiment using Retrieval Augmented Generation (RAG), with results shown in the right column. RAG is a method where source data from a corpus of information is shared with the LLM during generation, in the general framework of "Considering <INFORMATION CHUNKS> , answer the question <QUESTION>.". Herein, <INFORMATION CHUNKS> consists of chunks of data retrieved from the corpus of information, according to some measure of relevance for the question asked (this is realized in the work done here using a vector index that identifies relevant data using similarity measures via an embedding vector (details see **Materials and Methods**). In question answering without RAG, for comparison, we would directly ask the LLM for the answer, akin to: "Answer the question <QUESTION>.". In the first experiment, we observe similar responses. The conclusion of this experiment is that for cases that a LLM can handle on its own, the use of RAG does not significantly improve results.

We now move on to edge cases case where the original MechGPT model fails to produce accurate results, since we are inquiring about a highly specialized topic. We consider the case of a relatively newly synthesized material, molybdenene as reported in a recent paper [62] as an example. Since this paper was recently published, it was not included in the training data. When we ask MechGPT, the response is (here, question is in **bold** and response of the LLM in *italic font*):

> **What is molybdenene?**
>
> *Molybdenene is a proposed name for the theoretical monolayer of molybdenum atoms. It is not yet fabricated or observed in experiments.*

This response is wrong since this material has actually been fabricated (however, this paper was not included in the training set when the model was developed; it is also noted that sometimes a model may even hallucinate and produce made-up references). Such cases demonstrate one of several limitations of LLMs when used in a single-shot conventional generative paradigm. To address these issues, a user may decide to do a literature search or one can implement retrieval of live data automatically in the algorithm. As shown in **Figure 2**, a Google Scholar search identifies a recent paper that has produced this material. We then use this paper as data source for RAG. Here is an exchange with several question-answer pairs, showing much improved performance:

> **What is molybdenene?**
>
> *Molybdenene is a material that is synthesized from molybdenum. It forms a distinct square lattice structure composed of flat atomic sheets made up of molybdenum atoms. Molybdenene sheets have interatomic distances of 0.33-0.36 nm and exhibit staircase-like features. It can be grown in large areas and has been observed to have metallic Raman signatures. Molybdenene is different from its 2D oxides, such as MoO3 and MoO2, in terms of interatomic distance. It can be obtained in the form of whiskers consisting of weakly bonded molybdenene sheets. Molybdenene appears highly reflecting and flat under an optical microscope and can also be further processed to produce molybdenene oxide.*
>
> **Suggest ideas on how molybdenene could be used for electronic devices.**
>
> *Molybdenene, a two-dimensional material with high electrical conductivity and thermal stability, has potential applications in electronic devices. One idea is to use molybdenene as a channel material in field-effect transistors (FETs)* *due to its high carrier mobility. Another idea is to incorporate molybdenene into flexible electronics, taking advantage of its atomically thin structure that allows for flexibility and stretchability. Molybdenene could also be utilized in optoelectronic devices, such as photodetectors or light-emitting diodes (LEDs), due to its excellent light absorption and emission properties. Additionally, molybdenene's metallic nature and high carrier concentration make it suitable for use as electrodes in electronics and optoelectronics. It could also be used in molybdenene-based 2D-2D hybrids to couple quantum states at interfaces, which can play a crucial role in determining the properties of these hybrid materials. Overall, further research and development are needed to explore the full range of possibilities for molybdenene in electronic devices.*

**Suggest molecular modeling techniques suitable for molybdenene.**

*Molecular modeling techniques suitable for studying molybdenene include Density Functional Theory (DFT) calculations, molecular dynamics (MD) simulations, Monte Carlo simulations, force field-based simulations, and ab initio calculations. These techniques can provide insights into the molecular structure, electronic properties, mechanical properties, thermal behavior, and interactions of molybdenene with other molecules or materials. Raman spectroscopy can also be used to analyze the vibrational modes of molybdenene, complementing the molecular modeling studies.*

**Based on your general knowledge and the information provided in the context, what fracture behavior would you expect for molybdenene? Explain why.**

*Molybdenene is likely to exhibit brittle fracture behavior. This can be inferred from the context information that describes molybdenene as having a layered structure with weak interlayer bonding. The presence of staircase-like features and distinct interatomic distances further suggest a layered structure with weak bonding. Brittle fracture occurs with little or no plastic deformation, and cracks can easily propagate through the material, leading to sudden failure. The weak interlayer bonding in molybdenene would make it susceptible to brittle fracture, as cracks can easily propagate along the layers. Additionally, the observation of local strain-mediated structural evolution and the potential for crack branching and propagation in multiple directions indicate complex fracture behavior, which is often associated with brittle materials. Therefore, based on the information provided, it is reasonable to expect that molybdenene would exhibit brittle fracture behavior.*

These answers are generally accurate and agree with the findings reported in the original study [62]. Specific features about the structure (square lattice, four-fold symmetry dominance, staircase features) are accurately produced as can be confirmed by comparing against the original paper, as well as **Figure 2c-d** (reprinted images from the paper as relevant to the question asked). The model was also asked to suggest relevant modeling tools, as well as the anticipated fracture behavior. Key insights from these responses are that they are accurate based on current understanding. The predicted fracture behavior of "brittle" dominance is explained based on the weak bonding; however, the model also identifies potential complex fracture mechanisms due to deformation induced structural changes. Such structural changes under mechanical load were indeed identified in the original paper [62] and the model has accurately made a connection between this phenomenon and what might happen in a fracture scenario. Of course, more targeted studies of such a scenario is critical and we should view the LLM response as a direction that points to future hypothesis development that is then followed up by domain experts. In a scenario where a LLM may be used to aid such work, it may be tasked to develop a code and/or input file for a simulation that can then be executed with results interpreted further. In principle if relevant tools are implemented results from such augmented, new data, can be fed back into the knowledge basis and used for future generative tasks. This way, the LLM can see more context to make more accurate predictions. This workflow can be easily implemented in the current approach.

## 2.2 Interpretable graph-based strategies

Next, we explore a few other avenues by which the knowledge base is built and queried. In the initial experiment discussed above we used a vector indexing strategy based on text embeddings to match with a particular query, whereby Llama Index then assembles proper context for generation using the LLM. Another way to retain functional knowledge is the generation of Ontological Knowledge Graphs, which we find to typically provide better performance since it not only associates embeddings with a query but also develops an ontological understanding of if and how concepts are related. This endows the model with much deeper context when generating the response and allows it to answer more complex questions.

To start off we compare responses using conventional RAG and Ontological Knowledge Graphs. The question asked is:

> **Provide a summary of the transcending design principles across different materials classes. Provide examples and details.**

Here are two exchanges, based on reference [63]. First, the RAG method based on vector indexing and embeddings:

> *The transcending design principles across different materials classes involve the assembly of laminated and fibrillar structures, which define a higher level of organization of interfaces and nanostructures. These principles are observed in both bioinspired and synthetic inorganic-organic materials. The combination of strong and weak interfaces determines the toughness, strength, stiffness, shear, and adhesive properties of a material. Examples of materials that exhibit these design principles include bioinspired composites used in drug delivery, wearable electronics, human-computer interfaces, and structural components for the automotive and aerospace industries. These materials possess multifunctional attributes such as mechanical robustness, flexibility, transparency, sensing, adapting, optics, electrical conductivity, and thermal conductivity. The design principles also involve understanding the biogenesis or 'biofabrication' of materials, including the transition from fundamental building blocks to complex hierarchical structures. Digital manufacturing techniques, such as 3D printing, combined with advances in artificial intelligence, are also being used to create highly complex structures in materials design. The goal is to achieve properties such as optical clarity, loading with bioactive components, and tunable mechanical performance. Overall, the transcending design principles aim to create materials with enhanced properties and functionalities by integrating experiments with theory, modeling and simulation, data science, and artificial intelligence tools.*

Second, using Ontological Knowledge Graphs:

> *The transcending design principles across different materials classes involve the incorporation of hierarchical structures and multiple functionalities. These principles are observed in both natural and synthetic materials. Hierarchical structures play a crucial role in achieving desired mechanical properties in composite materials. They enable multiple deformation mechanisms, self-healing, plasticity, and toughening across different length scales. At the nanometer scale, compositional gradients and fuzzy interphases facilitate intrinsic toughening through chain slippage, stress delocalization, and non-destructive locking. At the micrometer scale, hierarchical structures contribute to extrinsic toughening through mechanisms like crack-bridging and energy dissipation via weak or soft interfaces.*
>
> *Multiple functionalities are achieved by incorporating diverse components and interfaces into synthetic processes. This involves combining different material phases, creating composition gradients, and enabling reversible energy dissipation. Examples of functionalities include shape morphing, stimuli-responsiveness, strengthening, toughening, stiffness, chirality, anisotropy, phase transitions, adhesion, self-healing, shear resistance, cracking resistance, and stress redirection.*
>
> *These design principles and functionalities are observed in various materials classes. For instance, fibrous composites found in nature, such as silk, keratin, and cartilage, exhibit hierarchical structures and multiple functionalities. In engineered heterogeneous materials, ceramic and ballistic armor utilize Bouligand structures for their optical, acoustic, and mechanical properties. Synthetic bioinspired analogues are being developed to replicate the stiffness, strength, and toughness found in natural materials.*
>
> *Overall, the transcending design principles involve the incorporation of hierarchical structures and multiple functionalities across different materials classes, both in nature and in synthetic materials. These principles enable materials to achieve desired mechanical properties and multiple functionalities, making them versatile and adaptable.*

The second example is more detailed and includes more mechanistic insights.

As an example to further explore Ontological Knowledge Graphs based generation, we choose a complex domain at the nexus of deep learning and flaw-tolerance that was discussed in recent literature [64]. **Table 2** depicts a comparison of responses of MechGPT alone, and MechGPT and Ontological Knowledge Graphs based RAG, for a specific topic around flaw tolerance in graphene. As can be seen MechGPT alone fails to provide accurate answers, similar as in the earlier example. When we use Ontological Knowledge Graphs built on the top cited result of a Google Scholar search in the topic of "Deep learning flaw tolerance" (Lew *et al*. [64]), results are much improved. As can be seen in **Table 2**, the model now accurately answers detailed questions about this scientific topic. This general approach also allows to inject new knowledge, or potentially new data/measurements or simulation results, to augment the base baked-in knowledge of the LLM and it is not limited to scientific reports, papers, patents, and similar but can also incorporate results from physics-based modeling (e.g., *ab-initio* or MD or FEM generated data). It demonstrates how the combination of more complex sampling strategies, potentially implemented via iterative strategies, can provide novel use cases that are particularly relevant for specific scientific scenarios.

We now focus more deeply on graph-based methods and specifically how they can be utilized effectively for mechanistic insights. **Figure 3** depicts details of the process that involves the construction of Ontological Knowledge Graphs from data sources These graphs are then used to provide context to generate responses for queries. **Figure 3a** shows an example based on a review paper on natural nanofibrils [63] and **Figure 3b** shows an example of a paper on fracture of intermediate filament networks, based on [65]. A few subgraphs are highlighted in the right panels to reveal the type of insights the model develops from the data provided, clearly showing the similarity in form to what is known as ontology logs [56,61,66,67]. It is noted that this information is used to provide useful context to a query – to help the model understand not only what concepts are important, but also how they are related. This additional feature of relatedness, expressed via directional graph edges, is uniquely added in the knowledge graph strategy, over a 'regular' RAG approach that relies on indexing based on embedding similarities. Specifically, when using graph-based RAG methods, we consider an additional feature of related data chunks derived from the directional graph edges. This means that we do not only collect chunks of relevant context from the vector indexing, but use the knowledge graph to understand how concepts, and associated data chunks, are related. This is done for up to two hops in the graph structure, thereby infusing concepts and their relationship to one another into the process of question answering. Results for these two specific papers are shown in **Tables 3** and **4**, respectively. The responses show excellent detail, a lucid tone, and provide relevant information.

We now move to explore interpretability for specific tasks executed by the LLM, not the entire knowledge graph. Indeed, we can directly visualize the task-specific Ontological Knowledge Graphs, here exemplified for the deformation behavior of protein filament networks (**Figure 4a,** as reported in [65]). For instance, **Figure 4b** shows a knowledge graph for the question:

> **Tell me more about flaw-tolerance in protein networks.**

for the case of the intermediate filament paper [65]. The extracted keywords: include 'tolerance', 'flaw', 'protein', 'networks', 'flaw-tolerance', 'protein networks'. From this a task-specific Ontological Knowledge Graph is constructed and visualized in **Figure 4b**, with additional context provided in **Table 5**. The model response is:

> *Flaw-tolerance in protein networks refers to their ability to withstand and tolerate structural irregularities or flaws without catastrophic failure. Unlike conventional solids, protein networks, particularly alpha-helical protein networks, can dissipate a significant amount of energy and stretch without affecting neighboring bonds. This is because there are no immediate interactions between individual filaments in the network that prevent microscopic rotations and shear, resulting in no strong stress concentrations at corners of cracks. The flaw-tolerance of protein networks is attributed to their unique geometric arrangement, the properties of individual protein domains, and the ability to dissipate energy and stretch without affecting neighboring bonds. Additionally, protein networks have a self-healing nature, as the utilization of hydrogen bonds allows the structure to reform at moderate temperatures and restore the initial alpha-helical structure even after severe deformation. This flaw-tolerance allows protein networks to withstand large deformations and stresses, making them highly resilient and mechanically robust. Understanding the mechanisms behind the flaw-tolerance of protein networks is important for developing models that describe the structure-property relationships in these networks under large deformation and for advancing knowledge in fields such as materials science and bioengineering.*

We now proceed to the response to the question:

> **What is the mechanism by which filaments ultimately fail?**

Extracted keywords are 'filaments', 'fail', and 'mechanism'. The response is:

> *The mechanism by which filaments ultimately fail is through the stretching of the alpha-helical protein domains, which leads to the breaking of H-bonds within and between the filaments. This stretching and breaking of bonds eventually causes the filaments to fail.*

The corresponding Ontological Knowledge Graphs is shown in **Figure 4c.** The graphs in **Figures 4b-c** provide detailed mechanistic insights into how the model recalls knowledge and answers the questions. A close inspection of these graphs can provide additional insights into the mechanisms of a system of interest, and thereby not only provide interpretability but also a potential to develop new hypothesis or research questions that can be realized in an AI-human collaboration.

Another is done using more sophisticated sampling methods that build on the concept of adapting the way we interact with a LLM and introducing alternatives to simple linear sampling, using techniques such as Chain-of-Thought prompting [12], Tree-of-Thought strategies [68] and related approaches [69]. **Figure 5** shows the basic mechanics of this approach, where **Figure 5a** shows the construction of responses based on direct, linear sampling versus more complex, multiple-step sampling (**Figure 5b**). As depicted in **Figure 5c**, the assessment of responses such as concepts using a LLM serving the role of a critic can provide powerful nonlinear sampling mechanisms that significantly enhance responses (and a multitude of ways can be implemented depending on use cases). In its simplest form, the general idea is that the LLM is first used to generate initial thoughts (and we can enhance the creativity of the model by increasing the sampling temperature – this may result in more explorative answers). Then, in a second step, the model is queried to develop the most critical concepts, and list them (these steps can be repeated multiple times). This set of critical concepts is then used to answer the original question, to yield the answer. This step-wise interactive use of LLMs can provide advantages to extract more accurate, more nuanced and more refined knowledge.

**Table 6** shows a specific example using multi-step tree sampling. The question asked is:

> **What would be likely failure mechanisms of a hybrid silk-metal nanocomposite, reinforced with carbon nanotubes?**

We proceed by asking several questions to sample dynamically. Compared to the single-shot answer, the complex prompting strategy elicits more nuanced and detailed responses. **Table 7** shows another result using multi-step tree sampling. Here, the question asked is:

> **The mechanism of flaw-tolerance in alpha-helical protein networks is: "The geometric transformation of the crack shape is believed to help minimize stress concentration at corners, although the exact mechanism is not fully understood. As the crack undergoes a series of changes, from mode I to a circular hole, and eventually becoming an elongated crack aligned with the direction of loading, the stress concentration at corners is reduced. This transformation occurs as individual proteins within the crack continuously unfold, independent of their neighboring proteins. The reduced stresses in the vicinity of the elongated crack contribute to a decrease in stress concentration at corners. Overall, this geometric transformation of the crack shape is thought to provide self-protection and distribute stress more evenly along the crack, ultimately minimizing stress concentration." Now, describe how to translate this concept to a material composed out of thin glass fibers as elementary building blocks. Define details of how the glass fibers must be engineered, and how they are arranged into larger-scale structures.**

We then proceed by asking several follow-up questions, with detailed results of the experiment provided via **Table 7**. Compared to the single-shot answer, the complex prompting strategy elicits more nuanced and detailed responses. This example showed how additional augmentation and context can be built into the query itself, to then trigger a rich set of generative steps with lots of nuanced responses that are helpful for engineering and scientific processes.

Finally, the opportunities of using LLMs offer interesting possibilities at the interface of knowledge domains, especially using Ontological Knowledge Graphs. **Figure 6** shows the results of such an experiment, where we use MechGPT to elucidate analogies between disparate domains, in this case between music and proteins. We use the MechGPT model to sample a set of responses to the query:

> **Discuss the analogies are between music and proteins. Specifically address counterpoint and how it relates music to protein. Provide a detailed and lengthy answer**.

Using GPT-4 we then assemble an Ontological Knowledge Graph from this set of data to elucidate key connections and relationships identified. We find that the key connector between the various subgraphs are "*basic units*", where these are amino acids in proteins and notes in music. The model further predicts several specific analogy concepts, such as that analogies "lie in the organization of units" and "help us understand wonders of life", among many others (subgraph in the right lower corner). Other connectors between subgraphs are identified as "structures" and "notes" as key nodes. This example can be expanded in future work, but it shows the capacity of using LLMs as a design tool to connect disparate domains of knowledge and to interact with data and knowledge in innovative ways that can augment conventional engineering workflows.

### 2.3 Multi-AI agent based strategies: Solving tasks using a team of LLMs

The previous examples were conducted by using one LLM to solve tasks ranging from question answering to graph generation. We now show how a set of LLM agents, organized into a virtual team of AIs (see **Figure 1d**), can effectively solve complex problems via collaborative interactions. The individual agents can have distinct properties, knowledge, and special powers. For instance, in the first example we create agents that have special powers to generate molecular coordinates from SMILES [70] representations of arbitrary molecules and also have powers to conduct a Density Function Theory (DFT) simulation [71]. The ability to conduct DFT simulations in particular is highly relevant since it provides access to new physics-based data; here, agents have the ability to complement baked-in knowledge from pretraining with new measurements or simulation results.

#### *2.3.1 Molecular design*

As a first example, **Table 8** shows a result of such an agent-based modeling, applied here to solve a molecular design problem. The question posed here is: "*Consider this molecule in SMILES representation: CCCCXCC, where X is one of (C, O, N). Which of these options leads to the lowest energy structure?*" The agents devise a plan, carry out the simulations, analyze the data and provide the final response to the User. The conversational transcript shown in the table provides a detailed account of how the problem is solved, and how the answer is ultimately provided. The set of agents includes the following entities:

- "User" (human that poses the question),
- "Planner" that develops a plan to answer the question,
- "Coordinate retriever" that can retrieve coordinates from SMILES coordinates or its own knowledge
- "Chatbot" with broad skills and execution power.

The agents have access to a Python function that converts SMILES code into atom types and molecular coordinates (the molecular structure), as well as a Python function that conducts a DFT simulation given a molecular structure as input. In the example provided here, the Python functions were provided by a human expert, but it is also possible that LLMs write and execute their own code. Having human experts code up such special functions aids to achieve a streamlined process by which key information is obtained. As can be seen in the table, the agents work out the solution and identify that CCCCOCC is the structure with the lowest energy. Such tasks can easily be extended to more complex design objectives including more complex molecular or other scenarios. Due to the use of custom functions, highly complex workflows can be implemented; where the LLM agents are aware of the various function calls they can execute (see **Materials and Methods** for details).

#### *2.3.2 Team of experts to answer complex research questions*

Agent-based models can also be used to answer complex research questions and inquiries using a team of LLMs with particularly organized knowledge via a set of expert agents. The set of agents include:

- "Boss" (human that poses the question),
- "Senior Engineer" that develops a plan to answer the question,
- "Modeling Expert"
- "Reviewer" who critically assesses the responses and integrates knowledge.

The boss is a proxy for the human user, who asks the questions and gives approval and input at different stages of the conversation; in this case we set up the conversation so that human input is sought periodically to allow for follow-up questions and refinements of the answer. Each of the agents can use experts with deep technical knowledge for additional information, and we define three expert agents using RAG with access to specific knowledge (see **Materials and Methods** for details). **Table 9** shows results of the agent set shown in **Figure 7,** used to answer a complex design question: "*Design molybdenene 2D materials by using ideas of flaw-tolerance in alpha-helical protein meshes. Develop three design principles that incorporate important mechanisms.*".

*2.3.3 Agent-based force field development: Coding, execution, data analysis, file saving and plotting*

The final example provides the most powerful illustration of how LLMs can not only answer complex questions via a team of agents but also write *de novo* code that is utilized to then develop an answer by executing it and collecting all relevant data via an active learning process by which physical ground truths are generated, via DFT modeling, on the fly. Sophisticated LLMs can also write and execute code, and need not rely in predetermined functions as in the previous example. **Figure 8** shows an experiment of using a set of two LLM agents, each with specific expertise and capabilities, to write, execute code, and analyze the data produced. Here, we exemplify this approach in automatic force field fitting. The task given is:

> **Use PySCF to calculate the energy of a O2 molecule, for varying bond lengths between 0.7 and 1.8 Angstroms in increments of 0.1 Angstroms. Fit a piecewise spline curve to the data. Then plot the energy over the bond length for the data and the fitted potential, and save as plot_O2_spline_fit_potential.svg. Save the parameters of the spline curve fit into a file, in JSON format. Save the calculation results in a file, in JSON format. What is the lowest energy configuration?**

Giving a general task, the set of agents automatically defines the problem and strategy, writes code, executes code, and provides the required out: A graph showing the raw energy vs. geometry data, JSON file outputs of energies, a spline fit to the data, including saving the force field parameters as a JSON file. The final output created by the set of two agents (**Figure 8a**) is:

> Great! The code executed successfully.
>
> The lowest energy configuration for the O2 molecule is at a bond length of approximately 1.3 Å with an energy of -148.08420987269093 Hartree.
>
> The plot of the energy over the bond length for the data and the fitted potential has been saved as 'plot_O2_spline_fit_potential.svg'.
>
> The parameters of the spline curve fit have been saved in a file named 'spline_params.json' in JSON format.
>
> The calculation results have been saved in a file named 'calculation_results.json' in JSON format.
>
> Please check these files in your current directory.

A full summary of the conversation between the agents is provided in **Table 10,** and **Figure 8b** shows key result of the experiments including the code produced, as well as the figure generated, and the JSON datafiles that feature a full representation of the raw DFT data and the force field fitting. We emphasize that no predetermined functioned are used. Instead, the LLMs write and execute all code simply driven by the task given. Such an AI system driven by an amalgamation of LLMs and other codes and methods is a modern realization of earlier concepts to develop materials design paradigms, such as the Computational Materials Design Facility (CMDF [49]). With the advent of increasingly sophisticated LLMs such tasks can be more easily realized at scale.

## 3. Conclusions

While fine-tuned LLMs are powerful tools, they can fall short to provide accurate response or to provide up-to-date scientific results. Through a series of strategies, especially using Retrieval-Augmented Generation strategies (RAG), the use of Ontological Knowledge Graphs, and sets of LLMs interacting as autonomous agents that collaborate to solve tasks, we showed that we can inject knowledge into the generative process to enhance model performance, adapt it to new data, and ultimately to being able to provide well-specified results. The case of Ontological Knowledge Graphs in particular is a powerful computational strategy to develop ontological insights akin to "ologs" as introduced in earlier literature on category theory [50–53] for use in materials science [38,54,56–58,61] (see, examples for such graphs pertinent to a particular scientific question in **Figure 4b-c**). Another avenue we discussed was the use of Tree-of-Thought sampling that overcomes the limitation of a single, linear sampling step to obtain results, and instead uses multi-directional and multi-dimensional sampling (**Figure 5**). Therein we use the LLM itself to critique its own responses to filter out the most relevant details. This takes advantage of the fact that LLMs are often excellent in providing first, generative results akin to brainstorming and also provide an accurate means to classify key responses. By breaking down a single task into multiple steps we can generate better, more informed and accurate outcomes.

We find that RAG and its variants are powerful methods that will likely play an important role in future evolutions of LLMs in engineering and science, especially when combined with nonlinear sampling methods. Other work in the use of retrieval-augmented LLM use cases such as mathematical proof development [72] underscore these exciting possibilities. In particular, the benefit of using Ontological Knowledge Graphs is that they provide interpretable results that human scientists can read, understand and use for either developing understanding, hypothesis or solve complex tasks such as the building block replacement problem [56]. The examples shown in the paper showcase the construction of such graphs not only on the general corpus of data (**Figure 3**) but also specifically to the answering mechanism of a question (**Figure 4b-c**) that helps to elucidate insights into how questions are responded to within the context of knowledge. This not only provides insights that can be used, say in an analysis or design process, but also generates detailed insights into the mechanics by which the LLM constructs an answer. This can help understand the origin of answers and can help to reverse engineer if a model predicts incorrect or unreliable results or if a user has questions.

Similar arguments can be made for complex nonlinear sampling strategies where the conventional linear mechanism of prompting and answer pairing is expanded to include interactive decision mechanisms, as also pointed out in [73], that are highly promising especially when combined with active, interactive or innovative data collection strategies (data collection can range from human experts to sensors to ab-initio modeling). These decision mechanisms can be automated by using the LLM to judge its own responses, or can be augmented by human expert input. This fundamentally changes the paradigm from a black-box retrieval system towards a more transparent tool that typically leads to more useful insights and can answer questions that cannot be reliably retrieved using single-shot answers [68,74,75]. In addition to the Ontological Knowledge Graphs constructed (**Figures 3-4**), **Table 2** is a key result of this paper, showing a comparison of responses of MechGPT alone, and with the use of RAG, for a specific topic around flaw tolerance in graphene. The RAG method uses the result of a Google Scholar search in the topic of "Deep learning flaw tolerance", downloading the most highly cited paper (by Lew *et al*. [64]). The result showed that MechGPT alone was not able to provide correct results for this specific, recently published paper but MechGPT with RAG was able to provide accurate results with specific details.

Another key advantage of using retrieval-augmented methods, and especially when combined with Ontological Knowledge Graphs methods is that sources used to generate responses can be identified. This information is included in the initial processing step that provides data about what document was used, and within the document, which section or paragraph was utilized. We can extract this information by examining metadata produced by the RAG algorithm or directly query the LLM to provide this detail. The combination of a fine-tuned special purpose LLM in conjunction with RAG or KG methods is hence not only more accurate, but also provides users with details on the origin and citation of concepts used. This provides a better integration with the scientific process and acknowledgement. Ontological Knowledge Graphs can also be automatically constructed to relate disparate concepts or domains of knowledge. As an example, we presented in **Figure 6** the results of using MechGPT to elucidate analogies between music and proteins, identifying a rich set of insights from key connecting principles, subgraphs, and an overall visual representation of this area of scholarship. It nicely builds on, and complements, earlier work [54,76–78] based on analytical or physics-based methods that attempted to generate such relationships. It also complements adversarial language-based strategies as proposed in [79] with a much different focus, specifically to capture natural language and its innate interpretability. Using the computational method proposed here, we can dive deeper into more complex cross-disciplinary connections. If the generated graph is further queried, we could also construct graphs specifically relevant to the answering mechanism of a question (akin to what we presented in **Figure 4b-c**). While this is left to future work, it is an exciting possibility to explore cross-cutting ties in an interpretable manner.

Finally, agent-based modeling is a powerful technique that offers enhanced problem-solving capacity, as shown in various examples including determining energetic details about molecular design and optimization (**Table 8**), question-answering via a team of AI agents (**Figure 7** and **Table 9**) and *de novo* code generation, execution and problem solving (here exemplified for force field development, see **Table 10** and **Figure 8**). The ability to generate code retrieve new physical data, and integrate various information for inclusion in easily readable file formats (like JSON) offers enhanced capabilities for computational materials design with an innate integrate of codes that operate across modalities and scales. The ability of the system of AI agents to develop code and processes (including error correction, as shown in **Table 10**) strategically through collaboration (including the possibility of human expert input) offers new paradigms for material design workflows.

Much future work is necessary to explore the behavior and use cases of LLMs and how they can be made relevant for engineering and science. We stress that more direct and simpler, and perhaps more effective, use cases of transformer models exist in various specific tasks such as property prediction and forward/inverse design tasks, as outlined in recent papers [15,30,59,80]. Another area of great relevance are various societal implications that can also relate to the way engineers work in the future [73,81]. So far, we deliberately focused much of the discussion and use cases of fine-tuned LLMs on open-source variants; however, other future work could explore the use of much larger LLMs such as GPT-3.5 or GPT-4, in general or fine-tuned variants, for tasks discussed in this study. This can provide additional insights into use cases where RAG and graph-based methods can be utilized. Careful assessment of the use of LLMs in a scientific and engineering context is critical, reflecting potentially harmful impacts such technologies can have if used carelessly. Further investigations may also focus on uses of larger open source models, such as the MechGPT-70b models based off the Llama-2-70b family of models (not used here). This, together with new larger-scale fine-tunable models such as GPT-3.5-turbo or GPT-4 can open even more powerful avenues for applications of LLMs in engineering, multiscale modeling and complex problem solving. The use of agent models is particularly useful also to infuse key physical insight; as done in the example we use first-principles modeling using DFT to enhance the AI system with novel data. This strategy can principally be used also to conduct various other operations such as physical soundness (e.g. mass conservation), whether a result satisfies one or more principles or a partial differential equation, or constraints or design objectives. The use of natural language, combined with capabilities of LLMs to effectively deal with numbers, data, code and its execution, offers a wide array of possibilities in materials science and engineering and beyond – including, but not limited to, interpretability and the possibility of direct engagement with human expert.

## 4. Materials and Methods

**Figure 1** provides a summary of the strategies used in this study. **Figure 1a** shows a general perspective of how LLMs, such as a fine-tuned model like MechGPT, use context and queries to provide an answer. In a conventional setting, LLMs are queried against their parameter-based knowledge that was developed during the initial training and/or fine-tuning stages. By developing a knowledge base from data sources as shown in **Figure 1b**, we can augment the response of an LLM by providing relevant context retrieved from the knowledge base with the question, to provide an answer (**Figure 1c**). Other mechanisms for sampling discussed in this paper include nonlinear sampling where we repeatedly query the LLM with a context and task, and iterate through multiple generations to ultimately develop an answer (see feedback loop from answer to formulate new context and query).

## 4.1 Initial fine-tuned MechGPT development

All work reported here is done based on fine-tuned LLMs derived from the Llama-2 architecture, an open LLM foundation model [46], developed for interactive chat modality. The original development of MechGPT is described in earlier papers [28,59]. We provide just a brief summary here for completeness. The MechGPT model is based on a pretrained LLM that is fine-tuned with 'textbook' mechanics and materials knowledge. In the first step, we use a general-purpose language model (in our case a combination of the Llama-2 70b chat model [38] and GPT-3.5/4 in order to extract knowledge from the text in the form of question-answer pairs. To do this we show the LLM sections of the source data (books, Wikipedia pages, etc.) and ask it to 1) develop a question to which the answer is the section considered, and then to 2) develop a concise summary of the section considered. This achieves multiple purposes, including a cleanup of the text; the question-answer pairs are then reviewed, cleaned up (remove any data with references to figures) and complemented by additional question-answer pairs curated from human feedback. The models are developed in PyTorch [82] and implemented within the Hugging Face ecosystem. Training of the MechGPT model is performed based off the Llama 2 transformer architecture, using the OpenOrca-Platypus2-13B [83] as basis (license for all Llama derivative works: https://github.com/facebookresearch/llama/blob/main/LICENSE). The extended dataset used for training the model includes ~2,600 question-answer pairs from the "*Atomistic modeling of materials failure*" textbook [60] and ~5,600 question-answer pairs developed from Wikipedia sources related to mechanics concepts with a total of ~8,200 question-answer pairs. The maximum length of question-answer pairs in the prompt template is 768 tokens. The prompt template, using "{Question}" and "{Answer}" as stand-ins for the corresponding question-answer pair, is:

> User: {Question}<|end_of_turn|>Assistant: {Answer}<|end_of_turn|>

For a specific example of a training pair of question and answer:

> User: How does hyperelasticity, or the elasticity of large strains, affect the dynamics of brittle fracture, and why is it important to consider this effect in understanding the behavior of materials under stress?<|end_of_turn|>Assistant: Hyperelasticity, or the elasticity of large strains, plays a crucial role in the dynamics of brittle fracture. Unlike linear elastic behavior, which assumes that materials will return to their original shape after stress, hyperelasticity takes into account the fact that materials can undergo large deformations without breaking. This means that the elastic behavior of materials under stress can be significantly different from the linear elastic behavior assumed in many existing theories of dynamic fracture.\n\nExperiments and computer simulations have shown that the crack propagation speed can be significantly reduced compared to the predictions by linear elastic theory. This discrepancy cannot be attributed solely to the presence of imperfections in real materials, as similar discrepancies are observed in molecular dynamics simulations of perfect atomic lattices.\n\nThe importance of considering hyperelasticity in understanding the behavior of materials under stress lies in its ability to explain phenomena related to crack branching and dynamic crack tip instability. Hyperelasticity can also help explain why the maximum crack propagation speed observed in some experiments and simulations is significantly lower than predicted by linear elastic theory.\n\nOverall, considering hyperelasticity is essential to developing a complete understanding of the dynamics of brittle fracture. By taking into account the large deformations that materials can undergo without breaking, researchers can better understand the behavior of materials under stress and develop more accurate predictions of their behavior.<|end_of_turn|>

Another one:

> User: How does molecular dynamics simulation differ from purely continuum mechanics theories in handling stress concentrations and describing bond breaking at the nanoscale?<|end_of_turn|>Assistant: Molecular dynamics simulations can handle stress concentrations and describe bond breaking at the nanoscale better than purely continuum mechanics theories. Unlike continuum mechanics, molecular dynamics simulations can accurately describe the behavior of materials at the atomic scale, where stress concentrations and bond breaking occur. This is because molecular dynamics simulations use simplistic model potentials to represent atomic interactions, allowing for a more straightforward and accurate description of the mechanics of brittle fracture at the nanoscale. Additionally, the use of simplistic model potentials enables immediate comparison with continuum theories, which can help develop a deeper understanding of the mechanics of brittle fracture at the nanoscale. Furthermore, advances in computational power allow for modeling at length scales on the order of micrometers, making atomistic-based modeling a promising tool for future research in the area of modeling nanomechanical phenomena and linking to continuum mechanical theories.<|end_of_turn|>

To fine-tune the model, we used a paged 32-bit AdamW optimizer ("paged_adamw_32bit") with a learning rate of LR=0.0002, ε =1E8, with a warmup ratio of 0.03, cosine learning rate schedule, and gradient norm clipping of 0.3. See **Figure S1** for a distribution of token length in the training set, as well as loss over steps, and learning rate (LR) over steps.

To parallelize training, we implemented the use of Hugging Face Accelerate (https://huggingface.co/docs/accelerate/index). The training objective used here is to maximize the likelihood of predicting the next token (that is, a letter, part of or a word) given the previous words, for the training set developed via question-answer pairs. We use Quantized Low-Rank Adaptation (QLoRA) [84,85] to fine-tune the model by adding additional trainable layers and freezing the original pretrained model to avoid catastrophic forgetting of the original knowledge base (this approach involves freezing parameters of the original pre-trained model and introducing small additional layers that consist of trainable rank decomposition matrices (added in each of the 40 transformer layers of the model), thereby significantly reducing the number of trainable parameters (LoRA rank=64, $LoRA_{\alpha}$=16, $LoRA_{dropout}$=0.1, modules with adaptor include "q_proj", "up_proj", "o_proj", "k_proj", "down_proj", "gate_proj", and "v_proj"). This reduction leads to improved memory efficiency and faster training throughput.

Once trained, we use the model for inference. Typical sampling temperatures are 0.4 (except for the initial thought collection in the tree search method, which uses a higher temperature to tend the output towards more creative responses).

### 4.2 Generative strategies and sampling methods

We use Llama Index (https://github.com/jerryjliu/llama_index) as a tool to implement Retrieval Augmented Generation, RAG, as well as the graph-based variant. RAG allows us to combine LLMs with an additional, new knowledge bases, live data or other external sources (see, schematics in **Figure 1b-c**). In this study we use RAG for knowledge-intensive natural language processing tasks by allowing LLMs to access data and incorporate new information. We use LangChain (https://github.com/langchain-ai/langchain) to implement embeddings via the "all-MiniLM-L6-v2" model (https://huggingface.co/sentence-transformers/all-MiniLM-L6-v2 and https://www.sbert.net/). The embedding model maps text sections to a 384 dimensional vector space that is then used by Llama Index for indexing.

For the graph-based methods, we further use NebulaGraph (https://github.com/vesoft-inc/nebula) in conjunction with Llama Index to conduct the experiments. NebulaGraph is an open-source, distributed graph database that can handle large-scale graphs with high efficiency. In the context of Llama Index it provides useful features that allow the extraction of Ontological Knowledge Graphs for specific contexts and questions asked, and the ability to visualize graphs using Networkx (https://networkx.org/). We use GPT-3.5-turbo and GPT-4 to construct Ontological Knowledge Graphs from the raw data due to computational efficiency, and the fact that a general-purpose model is suitable for this initial extraction step (principally this can be done with any other model).

Multi-step sampling inspired by the Tree-of-Thought method is implemented by successively sampling the model using the series of questions delineated in **Figure 5b**. The series of interactions with the model unfolds using the following prompts to answer '{question}':

1. Generate a list of initial thoughts that are relevant for answering this question: '{question}'. Do not answer the question.
2. Read this: '{Response #1}'. List the most important concepts to answer the question '{question}'.
3. Considering '{Response #2}', answer this question with a detailed response: {question}

It is noted that both, nonlinear sampling and RAG can help to minimize hallucinations. Nonlinear sampling tends to be more powerful in larger models; RAG is effective even in small models.

### 4.3 Agent-based modeling: Autonomous interactions of multiple LLMs to solve complex multimodal problems

Agent modeling is implemented using AutoGen as the overarching framework (https://github.com/microsoft/autogen) [86]. This framework provides access to a Python implementation of concurrent integration of text generation, prompting, and logical control that allows us to exploit the natural text, code and processing modality of LLMs, including the possibility to execute code. We consider a set of agents with different behaviors, properties, and capabilities (details in text, but an overview can be seen in **Figure 1d**). The agents then conduct a collaborative conversation to solve a task. The conversation may or may not involve human input (if human input is prompted, the user can ask follow-up questions, clarify issues, or give further guidance to the AI system). All agents use GPT-4 as a backbone LLM.

Three example setups are used, with agents that have different capabilities. In the first example (Section 2.3.1) we define four agents: "User", "Planner", "Coordinate retriever" and "Chatbot" (details in text). To facilitate the specific tasks of coordinate generation from SMILES code, we define a function "coords_from_SMILES" that is made available to the agents, with the description "With a SMILES string as input, provides atom type and coordinates of a molecule." We use RDKit (https://github.com/rdkit/rdkit) for the conversion of SMILEs codes to coordinates (we first convert the SMILES codes to coordinates, then add H atoms, then relax the structure using MMFF94 [87]. Further, a function is defined to conduct a DFT simulation, "query_DFT", with the description: "With coordinates as input, calculate the energy of a molecule." The Density Functional Theory simulations are carried out using PySCF (https://pyscf.org/) [88–90]. This code was chosen since it is easily executable within a Python environment (in general, any other simulation code can be used).

The Coordinate retriever is instructed, via its system message, to retrieve coordinates in a certain format:

> You retrieve coordinates of molecules from SMILES strings or based on your own knowledge. The coordinates of the molecule must be provided in the following form, in units of Angstrom:
>
> coordinates = 'C 0.000000    0.000000    0.117790;C        0.000000    0.000000    0.117790'

The "Planner" has the following system message:

> Planner. Suggest a plan to solve the task.

The "Chatbot" has the following system message:

> You carry out energy calculations, and answer the task. Reply TERMINATE when the task is done.

Finally, the agent "User" that acts on behalf of the human user, is instructed as follows:

> You interact with the planner to develop a plan to solve the problem.

The "TERMINATE" message is used to signal that the question has been answered; once received the conversation ends. **Tables S1-3** provide code snippets to showcase the definition of this setup.

In the second example (section 2.3.2), we use a total of 7 agents, as defined in **Figure 7**. The system message of the "Boss" is:

> Boss who asks questions and gives tasks. Interact with the planner to approve the plan. Reply `TERMINATE` in the end when everything is done.

We further instruct the "Boss" to request human input when the "TERMINATE" message is received; that is, before the conversation is finalized the human user has the opportunity to either end the conversation or ask a follow-up question. The "Senior Engineer" is instructed as follows:

> Planner. You are a senior materials scientist with broad knowledge. Suggest a plan. Revise the plan based on feedback from the researcher, modeling expert, and reviewer, and ask the boss for approval. The plan may involve a researcher who can retrieve information about materials, a modeling expert who suggests modeling methods, and a reviewer who gives critical feedback. Explain the plan first. Be clear which step is performed by the researcher, which step is performed by the modeling expert, and which step is performed by the reviewer. Once the plan is created, ask the boss to approve.

The “Modeling expert” is instructed as follows:

> Modeling expert. You follow an approved plan. You are an expert in atomistic and multiscale modeling who contributes to the discussion by providing ideas for how simulation can enhance solving the problem.

The reviewer is instructed:

> Reviewer. You follow an approved plan. You are a scientific reviewer who gives critical feedback, adds new ideas, and integrates the concepts.

The expert agents defined in **Figure 7** use RAG to generate responses; where each of the agents (agents #5, #6, and #7 use data generated from the raw PDF of the respective papers, converted into markup language using the Nougat [91] optical character recognition (OCR) model. The system messages for the protein, molybdenene, and modeling experts are, respectively: “Assistant who has extra content retrieval power for solving difficult problems in protein materials.”, “Assistant who has extra content retrieval power for solving difficult problems in molybdenene materials, and other 2D materials.” and “Assistant who has extra content retrieval power for information related to atomistic and multiscale modeling, especially related to materials failure.” We use Chroma to develop the vector index (https://github.com/chroma-core/chroma) with all-MiniLM-L6-v2 as embedding model, as above.

For the results shown in Section 2.3.3 no predetermined Python functions are used. Instead, the LLMs write and execute all code simply driven by the task given (code generated is listed in the results table, including iterative error correction, and in **Figure 8**). To do this we define two agents, an “Assistant” that helps develop code and answer questions, and a “User” that is a proxy for the human that can autonomously execute code. Code execution is conducted directly in the local environment (optionally, code execution can also be done in a Docker container, which works in the same way but can be a bit slower since Docker images are downloaded and launched at every code execution step). As can be seen in **Table 10**, the agents not only write and execute code but also develop a plan to install proper Python libraries needed for the execution. Note, we specify a folder name in which the codes generated by the agents are stored so that they can be retrieved separately and used for other purposes.

We emphasize that for the results in **Table 8, Table 9** and **Figure 7** we use specially designed functions to i) retrieve 3D coordinates from SMILES codes and ii) conduct a DFT simulation based on coordinates. In the results in **Figure 8** and **Table 10,** no such functions are necessary since we reply on the LLM’s capability to write code/input files, and conduct a variety of other tasks.

## Supplementary Information

Additional Supplementary Information is provided, including additional figures and codes. Codes, model weights, the dataset and other materials can be found at https://github.com/lamm-mit/MeLM.

## Acknowledgements

This work was supported by MIT’s Generative AI initiative, Google, the Army Research Office (W911NF1920098 & W911NF2220213), ONR (N00014-19-1-2375 and N00014-20-1-2189), USDA (2021-69012-35978), as well as NIH U01. Additional support was provided by the MIT-IBM Watson AI Lab.

## References

[1] A. Vaswani, N. Shazeer, N. Parmar, J. Uszkoreit, L. Jones, A.N. Gomez, Ł. Kaiser, I. Polosukhin, Attention is all you need, in: Adv Neural Inf Process Syst, Neural information processing systems foundation, 2017: pp. 5999–6009. https://arxiv.org/abs/1706.03762v5 (accessed June 27, 2021).

[2] T.B. Brown, B. Mann, N. Ryder, M. Subbiah, J. Kaplan, P. Dhariwal, A. Neelakantan, P. Shyam, G. Sastry, A. Askell, S. Agarwal, A. Herbert-Voss, G. Krueger, T. Henighan, R. Child, A. Ramesh, D.M. Ziegler, J. Wu, C. Winter, C. Hesse, M. Chen, E. Sigler, M. Litwin, S. Gray, B. Chess, J. Clark, C.

Berner, S. McCandlish, A. Radford, I. Sutskever, D. Amodei, Language Models are Few-Shot Learners, Adv Neural Inf Process Syst. 2020-December (2020). https://arxiv.org/abs/2005.14165v4 (accessed June 26, 2023).

[3] A. Chowdhery, S. Narang, J. Devlin, M. Bosma, G. Mishra, A. Roberts, P. Barham, H.W. Chung, C. Sutton, S. Gehrmann, P. Schuh, K. Shi, S. Tsvyashchenko, J. Maynez, A. Rao, P. Barnes, Y. Tay, N. Shazeer, V. Prabhakaran, E. Reif, N. Du, B. Hutchinson, R. Pope, J. Bradbury, J. Austin, M. Isard, G. Gur-Ari, P. Yin, T. Duke, A. Levskaya, S. Ghemawat, S. Dev, H. Michalewski, X. Garcia, V. Misra, K. Robinson, L. Fedus, D. Zhou, D. Ippolito, D. Luan, H. Lim, B. Zoph, A. Spiridonov, R. Sepassi, D. Dohan, S. Agrawal, M. Omernick, A.M. Dai, T.S. Pillai, M. Pellat, A. Lewkowycz, E. Moreira, R. Child, O. Polozov, K. Lee, Z. Zhou, X. Wang, B. Saeta, M. Diaz, O. Firat, M. Catasta, J. Wei, K. Meier-Hellstern, D. Eck, J. Dean, S. Petrov, N. Fiedel, PaLM: Scaling Language Modeling with Pathways, (2022). http://arxiv.org/abs/2204.02311 (accessed September 23, 2023).

[4] R. Taylor, M. Kardas, G. Cucurull, T. Scialom, A. Hartshorn, E. Saravia, A. Poulton, V. Kerkez, R. Stojnic, Galactica: A Large Language Model for Science, (2022). http://arxiv.org/abs/2211.09085 (accessed September 23, 2023).

[5] Y. Ge, W. Hua, K. Mei, J. Ji, J. Tan, S. Xu, Z. Li, Y. Zhang, OpenAGI: When LLM Meets Domain Experts, (2023). http://arxiv.org/abs/2304.04370 (accessed September 23, 2023).

[6] OpenAI, GPT-4 Technical Report, (2023). http://arxiv.org/abs/2303.08774 (accessed September 23, 2023).

[7] S. Bubeck, V. Chandrasekaran, R. Eldan, J. Gehrke, E. Horvitz, E. Kamar, P. Lee, Y.T. Lee, Y. Li, S. Lundberg, H. Nori, H. Palangi, M.T. Ribeiro, Y. Zhang, Sparks of Artificial General Intelligence: Early experiments with GPT-4, (2023). https://arxiv.org/abs/2303.12712v5 (accessed April 23, 2023).

[8] R. Nadkarni, D. Wadden, I. Beltagy, N.A. Smith, H. Hajishirzi, T. Hope, Scientific Language Models for Biomedical Knowledge Base Completion: An Empirical Study, (2021). http://arxiv.org/abs/2106.09700 (accessed September 23, 2023).

[9] I. Beltagy, K. Lo, A. Cohan, SCIBERT: A pretrained language model for scientific text, EMNLP-IJCNLP 2019 - 2019 Conference on Empirical Methods in Natural Language Processing and 9th International Joint Conference on Natural Language Processing, Proceedings of the Conference. (2019) 3615–3620. https://doi.org/10.18653/v1/d19-1371.

[10] T. Schick, J. Dwivedi-Yu, R. Dessì, R. Raileanu, M. Lomeli, L. Zettlemoyer, N. Cancedda, T. Scialom, Toolformer: Language Models Can Teach Themselves to Use Tools, (2023). http://arxiv.org/abs/2302.04761 (accessed September 23, 2023).

[11] G. Mialon, R. Dessì, M. Lomeli, C. Nalmpantis, R. Pasunuru, R. Raileanu, B. Rozière, T. Schick, J. Dwivedi-Yu, A. Celikyilmaz, E. Grave, Y. LeCun, T. Scialom, Augmented Language Models: a Survey, (2023). http://arxiv.org/abs/2302.07842 (accessed September 23, 2023).

[12] J. Wei, X. Wang, D. Schuurmans, M. Bosma, B. Ichter, F. Xia, E. Chi, Q. Le, D. Zhou, Chain-of-Thought Prompting Elicits Reasoning in Large Language Models, (2022). http://arxiv.org/abs/2201.11903 (accessed September 23, 2023).

[13] N.R. Brodnik, S. Carton, C. Muir, S. Ghosh, D. Downey, M.P. Echlin, T.M. Pollock, S. Daly, Perspective: Large Language Models in Applied Mechanics, J Appl Mech. 90 (2023) 1–12. https://doi.org/10.1115/1.4062773.

[14] M.J. Buehler, Generative pretrained autoregressive transformer graph neural network applied to the analysis and discovery of novel proteins, J Appl Phys. 134 (2023) 84902. https://doi.org/10.1063/5.0157367.

[15] R.K. Luu, M. Wysokowski, M.J. Buehler, Generative Discovery of Novel Chemical Designs using Diffusion Modeling and Transformer Deep Neural Networks with Application to Deep Eutectic Solvents, Appl Phys Lett. 122 (2023). https://doi.org/10.1063/5.0155890.

[16] Y. Hu, M.J. Buehler, Deep language models for interpretative and predictive materials science, APL Machine Learning. 1 (2023) 010901. https://doi.org/10.1063/5.0134317.

[17] R. Azamfirei, S.R. Kudchadkar, J. Fackler, Large language models and the perils of their hallucinations, Crit Care. 27 (2023). https://doi.org/10.1186/S13054-023-04393-X.

[18] N. Kandpal, H. Deng, A. Roberts, E. Wallace, C. Raffel, Large Language Models Struggle to Learn Long-Tail Knowledge, (2022). http://arxiv.org/abs/2211.08411 (accessed September 23, 2023).

[19] N. Varshney, W. Yao, H. Zhang, J. Chen, D. Yu, A Stitch in Time Saves Nine: Detecting and Mitigating Hallucinations of LLMs by Validating Low-Confidence Generation, (2023). https://arxiv.org/abs/2307.03987v2 (accessed September 23, 2023).

[20] Z. Ji, N. Lee, R. Frieske, T. Yu, D. Su, Y. Xu, E. Ishii, Y.J. Bang, A. Madotto, P. Fung, Survey of Hallucination in Natural Language Generation, ACM Comput Surv. 55 (2023). https://doi.org/10.1145/3571730.

[21] N. McKenna, T. Li, L. Cheng, M.J. Hosseini, M. Johnson, M. Steedman, Sources of Hallucination by Large Language Models on Inference Tasks, (2023). https://arxiv.org/abs/2305.14552v1 (accessed September 23, 2023).

[22] Y. Ge, W. Hua, K. Mei, J. Ji, J. Tan, S. Xu, Z. Li, Y. Zhang, OpenAGI: When LLM Meets Domain Experts, (2023). https://arxiv.org/abs/2304.04370v5 (accessed September 23, 2023).

[23] S. Harrer, Attention is not all you need: the complicated case of ethically using large language models in healthcare and medicine, EBioMedicine. 90 (2023). https://doi.org/10.1016/J.EBIOM.2023.104512.

[24] Y. Liang, R. Zhang, L. Zhang, P. Xie, DrugChat: Towards Enabling ChatGPT-Like Capabilities on Drug Molecule Graphs, (2023). https://arxiv.org/abs/2309.03907v1 (accessed September 12, 2023).

[25] S. Wolfram, What is ChatGPT doing ... and why does it work?, 2023, https://writings.stephenwolfram.com/2023/02/what-is-chatgpt-doing-and-why-does-it-work/.

[26] R.K. Luu, M.J. Buehler, Materials Informatics Tools in the Context of Bio-Inspired Material Mechanics, J Appl Mech. 90 (2023). https://doi.org/10.1115/1.4062310.

[27] W. Lu, N.A. Lee, M.J. Buehler, Modeling and design of heterogeneous hierarchical bioinspired spider web structures using deep learning and additive manufacturing, Proceedings of the National Academy of Sciences. 120 (2023) e2305273120. https://doi.org/10.1073/PNAS.2305273120.

[28] R.K. Luu, M.J. Buehler, BioinspiredLLM: Conversational Large Language Model for the Mechanics of Biological and Bio-inspired Materials, (2023). https://arxiv.org/abs/2309.08788v1 (accessed September 23, 2023).

[29] W. Lu, D.L. Kaplan, M.J. Buehler, Generative modeling, design and analysis of spider silk protein sequences for enhanced mechanical properties, (2023). https://arxiv.org/abs/2309.10170v1 (accessed September 23, 2023).

[30] M.J. Buehler, MeLM, a generative pretrained language modeling framework that solves forward and inverse mechanics problems, J Mech Phys Solids. (2023) 105454. https://doi.org/10.1016/J.JMPS.2023.105454.

[31] J. Jumper, R. Evans, A. Pritzel, T. Green, M. Figurnov, O. Ronneberger, K. Tunyasuvunakool, R. Bates, A. Žídek, A. Potapenko, A. Bridgland, C. Meyer, S.A.A. Kohl, A.J. Ballard, A. Cowie, B. Romera-Paredes, S. Nikolov, R. Jain, J. Adler, T. Back, S. Petersen, D. Reiman, E. Clancy, M. Zielinski, M. Steinegger, M. Pacholska, T. Berghammer, S. Bodenstein, D. Silver, O. Vinyals, A.W. Senior, K. Kavukcuoglu, P. Kohli, D. Hassabis, Highly accurate protein structure prediction with AlphaFold, Nature. 596 (2021) 583–589. https://doi.org/10.1038/s41586-021-03819-2.

[32] F.Y.C. Liu, B. Ni, M.J. Buehler, PRESTO: Rapid protein mechanical strength prediction with an end-to-end deep learning model, Extreme Mech Lett. 55 (2022) 101803. https://doi.org/10.1016/J.EML.2022.101803.

[33] C.H. Yu, W. Chen, Y.H. Chiang, K. Guo, Z. Martin Moldes, D.L. Kaplan, M.J. Buehler, End-to-End Deep Learning Model to Predict and Design Secondary Structure Content of Structural Proteins, ACS Biomater Sci Eng. 8 (2022) 1156–1165. https://doi.org/10.1021/ACSBIOMATERIALS.1C01343.

[34] Y. Hu, M.J. Buehler, End-to-End Protein Normal Mode Frequency Predictions Using Language and Graph Models and Application to Sonification, ACS Nano. 16 (2022) 20656–20670. https://doi.org/10.1021/ACSNANO.2C07681/SUPPL_FILE/NN2C07681_SI_004.PDF.

[35] C.H. Yu, E. Khare, O.P. Narayan, R. Parker, D.L. Kaplan, M.J. Buehler, ColGen: An end-to-end deep learning model to predict thermal stability of de novo collagen sequences, J Mech Behav Biomed Mater. 125 (2022) 104921. https://doi.org/10.1016/J.JMBBM.2021.104921.

[36] E. Khare, C. Gonzalez-Obeso, D.L. Kaplan, M.J. Buehler, CollagenTransformer: End-to-End Transformer Model to Predict Thermal Stability of Collagen Triple Helices Using an NLP Approach, ACS Biomater Sci Eng. 2022 (2022) 4310. https://doi.org/10.1021/ACSBIOMATERIALS.2C00737/ASSET/IMAGES/MEDIUM/AB2C00737_M001.GIF.

[37] E.L. Buehler, M.J. Buehler, End-to-end prediction of multimaterial stress fields and fracture patterns using cycle-consistent adversarial and transformer neural networks, Biomedical Engineering Advances. 4 (2022) 100038. https://doi.org/10.1016/J.BEA.2022.100038.

[38] M.J. Buehler, FieldPerceiver: Domain agnostic transformer model to predict multiscale physical fields and nonlinear material properties through neural ologs, Materials Today. 57 (2022) 9–25. https://doi.org/10.1016/J.MATTOD.2022.05.020.

[39] C. Yang, X. Wang, Y. Lu, H. Liu, Q. V. Le, D. Zhou, X. Chen, Large Language Models as Optimizers, (2023). https://arxiv.org/abs/2309.03409v1 (accessed September 12, 2023).

[40] A.S. Fuhr, B.G. Sumpter, Deep Generative Models for Materials Discovery and Machine Learning-Accelerated Innovation, Front Mater. 9 (2022) 865270. https://doi.org/10.3389/FMATS.2022.865270/BIBTEX.

[41] B. Ni, D.L. Kaplan, M.J. Buehler, Generative design of de novo proteins based on secondary structure constraints using an attention-based diffusion model, Chem. (2023) https://doi.org/10.1016/j.chempr.2023.03.020. https://doi.org/https://doi.org/10.1016/j.chempr.2023.03.020.

[42] P. Inguva, V.J. Bhute, T.N.H. Cheng, P.J. Walker, Introducing students to research codes: A short course on solving partial differential equations in Python, Education for Chemical Engineers. 36 (2021) 1–11. https://doi.org/10.1016/J.ECE.2021.01.011.

[43] S. Milano, J.A. McGrane, S. Leonelli, Large language models challenge the future of higher education, Nature Machine Intelligence 2023 5:4. 5 (2023) 333–334. https://doi.org/10.1038/s42256-023-00644-2.

[44] W.M. Lim, A. Gunasekara, J.L. Pallant, J.I. Pallant, E. Pechenkina, Generative AI and the future of education: Ragnarök or reformation? A paradoxical perspective from management educators, The International Journal of Management Education. 21 (2023) 100790. https://doi.org/10.1016/J.IJME.2023.100790.

[45] OpenAI, GPT-4 Technical Report, (2023). https://arxiv.org/abs/2303.08774v3 (accessed September 23, 2023). This is a website, no authors provided (to be cited as OpenAI (2023).

[46] H. Touvron, L. Martin, K. Stone, P. Albert, A. Almahairi, Y. Babaei, N. Bashlykov, S. Batra, P. Bhargava, S. Bhosale, D. Bikel, L. Blecher, C.C. Ferrer, M. Chen, G. Cucurull, D. Esiobu, J. Fernandes, J. Fu, W. Fu, B. Fuller, C. Gao, V. Goswami, N. Goyal, A. Hartshorn, S. Hosseini, R. Hou, H. Inan, M. Kardas, V. Kerkez, M. Khabsa, I. Kloumann, A. Korenev, S. Koura, M.-A. Lachaux, T. Lavril, J. Lee, D. Liskovich, Y. Lu, Y. Mao, X. Martinet, T. Mihaylov, P. Mishra, I. Molybog, Y. Nie, A. Poulton, J. Reizenstein, R. Rungta, K. Saladi, A. Schelten, R. Silva, E. Michael, S. Ranjan, S. Xiaoqing, E. Tan, B. Tang, R. Taylor, A. Williams, J.X. Kuan, P. Xu, Z. Yan, I. Zarov, Y. Zhang, A. Fan, M. Kambadur, S. Narang, A. Rodriguez, R. Stojnic, S. Edunov, T. Scialom, Llama 2: Open Foundation and Fine-Tuned Chat Models, (2023). https://arxiv.org/abs/2307.09288v2 (accessed August 22, 2023).

[47] Falcon LLM, https://falconllm.tii.ae/ (accessed June 26, 2023).

[48] G. Penedo, Q. Malartic, D. Hesslow, R. Cojocaru, A. Cappelli, H. Alobeidli, B. Pannier, E. Almazrouei, J. Launay, The RefinedWeb Dataset for Falcon LLM: Outperforming Curated Corpora with Web Data, and Web Data Only, (2023). https://arxiv.org/abs/2306.01116v1 (accessed June 26, 2023).

[49] M.J. Buehler, J. Dodson, A.C.T. Van Duin, P. Meulbroek, W.A. Goddard, The Computational Materials Design Facility (CMDF): A powerful framework for multi-paradigm multi-scale simulations, MRS Online Proceedings Library (OPL). 894 (2005) 0894-LL03-03. https://doi.org/10.1557/PROC-0894-LL03-03.

[50] S. Eilenberg, S. Mac Lane, General theory of natural equivalences, Trans Am Math Soc. 58 (1945) 247. https://doi.org/10.1090/S0002-9947-1945-0013131-6.

[51] S. Eilenberg, S. MacLane, Group Extensions and Homology, Ann Math. 43 (1942) 757–831. https://doi.org/10.2307/1968966.

[52] J.-P. Marquis, Category Theory, Stanford Encyclopedia of Philosophy. (2019). https://plato.stanford.edu/entries/category-theory/ (accessed September 23, 2023).

[53] J.C. Baez, M. Stay, New Structures in Physics, Lecture Notes in Physics, Vol. 813. (2011) 95–174. http://math.ucr.edu/home/baez/rosetta/rose3.pdf (accessed September 23, 2023).

[54] T. Giesa, D.I. Spivak, M.J. Buehler, Reoccurring Patterns in Hierarchical Protein Materials and Music: The Power of Analogies, Bionanoscience. 1 (2011). https://doi.org/10.1007/s12668-011-0022-5.

[55] D.I. Spivak, M.J.B. Reoccurring, Reoccurring Patterns in Hierarchical Protein R eoccurring patterns in hierarchical protein materials and music : The power of analogies, (2011) 0–13.

[56] T. Giesa, D.I. Spivak, M.J. Buehler, Category theory based solution for the building block replacement problem in materials design, Adv Eng Mater. 14 (2012). https://doi.org/10.1002/adem.201200109.

[57] T. Giesa, R. Jagadeesan, D.I. Spivak, M.J. Buehler, Matriarch: A Python Library for Materials Architecture, ACS Biomater Sci Eng. 1 (2015). https://doi.org/10.1021/acsbiomaterials.5b00251.

[58] D.B. Brommer, T. Giesa, D.I. Spivak, M.J. Buehler, Categorical prototyping: Incorporating molecular mechanisms into 3D printing, Nanotechnology. 27 (2015). https://doi.org/10.1088/0957-4484/27/2/024002.

[59] M.J. Buehler, MechGPT, a language-based strategy for mechanics and materials modeling that connects knowledge across scales, disciplines and modalities, Appl. Mech. Rev. (2023). https://doi.org/10.1115/1.4063843 (accessed October 18, 2023).

[60] M.J. Buehler, Atomistic modeling of materials failure, 2008. https://doi.org/10.1007/978-0-387-76426-9.

[61] D.I. Spivak, T. Giesa, E. Wood, M.J. Buehler, Category theoretic analysis of hierarchical protein materials and social networks, PLoS One. 6 (2011). https://doi.org/10.1371/journal.pone.0023911.

[62] T.K. Sahu, N. Kumar, S. Chahal, R. Jana, S. Paul, M. Mukherjee, A.H. Tavabi, A. Datta, R.E. Dunin-Borkowski, I. Valov, A. Nayak, P. Kumar, Microwave synthesis of molybdenene from MoS2, Nature Nanotechnology 2023. (2023) 1–9. https://doi.org/10.1038/s41565-023-01484-2.

[63] S. Ling, D.L. Kaplan, M.J. Buehler, Nanofibrils in nature and materials engineering, Nat Rev Mater. 3 (2018). https://doi.org/10.1038/natrevmats.2018.16.

[64] A.J. Lew, C. Yu, Y. Hsu, M.J. Buehler, Deep learning model to predict fracture mechanisms of graphene, npj 2D materials, 2021, Vo. 5(48). https://www.nature.com/articles/s41699-021-00228-x

[65] T. Ackbarow, D. Sen, C. Thaulow, M.J. Buehler, Alpha-helical protein networks are self-protective and flaw-tolerant, PLoS One. 4 (2009). https://doi.org/10.1371/journal.pone.0006015.

[66] S.W. Cranford, M.J. Buehler, Biomateriomics, Springer Netherlands, 2012.

[67] The Future of Biomateriomics, (2012) 425–430. https://doi.org/10.1007/978-94-007-1611-7.

[68] S. Yao, D. Yu, G. Deepmind, J. Zhao, T.L. Griffiths, Y. Cao, K. Narasimhan, Tree of Thoughts: Deliberate Problem Solving with Large Language Models, (2023). https://arxiv.org/abs/2305.10601v1 (accessed September 23, 2023).

[69] S. Dhuliawala, M. Ai, E. Zürich, M. Komeili, J. Xu, R. Raileanu, X. Li, A. Celikyilmaz, J. Weston, Chain-of-Verification Reduces Hallucination in Large Language Models, (2023). https://arxiv.org/abs/2309.11495v1 (accessed September 24, 2023).

[70] D. Weininger, SMILES, a Chemical Language and Information System: 1: Introduction to Methodology and Encoding Rules, J Chem Inf Comput Sci. 28 (1988) 31–36. https://doi.org/10.1021/CI00057A005/ASSET/CI00057A005.FP.PNG_V03.

[71] R.G. Parr, Y. Weitao, Density-Functional Theory of Atoms and Molecules, Density-Functional Theory of Atoms and Molecules. (1995). https://doi.org/10.1093/OSO/9780195092769.001.0001.

[72] K. Yang, A.M. Swope, A. Gu, R. Chalamala, P. Song, S. Yu, S. Godil, R. Prenger, A. Anandkumar, U. Santa Barbara, U. Austin, LeanDojo: Theorem Proving with Retrieval-Augmented Language Models, (2023). https://arxiv.org/abs/2306.15626v1 (accessed September 23, 2023).

[73] T. van der Zant, M. Kouw, L. Schomaker, Generative artificial intelligence, Studies in Applied Philosophy, Epistemology and Rational Ethics. 5 (2013) 107–120. https://doi.org/10.1007/978-3-642-31674-6_8/COVER.

[74] J. Long, Large Language Model Guided Tree-of-Thought, (2023). https://arxiv.org/abs/2305.08291v1 (accessed September 23, 2023).

[75] W. Chen, X. Ma, X. Wang, W.W. Cohen, Program of Thoughts Prompting: Disentangling Computation from Reasoning for Numerical Reasoning Tasks, (2022). https://arxiv.org/abs/2211.12588v3 (accessed September 23, 2023).

[76] S.L. Franjou, M. Milazzo, C.-H. Yu, M.J. Buehler, A perspective on musical representations of folded protein nanostructures, Nano Futures. 5 (2021) 12501.

[77] C.H. Yu, Z. Qin, F. Martin-Martinez, M.J. Buehler, A self-consistent sonification method to translate amino acid sequences into musical compositions and application in protein design using AI, ACS Nano. 13 (2019) 7471–7482.

[78] M. Milazzo, G.I. Anderson, M.J. Buehler, Bioinspired translation of classical music into de novo protein structures using deep learning and molecular modeling, Bioinspir Biomim. 17, 015001, DOI 10.1088/1748-3190/ac338a

[79] M.J. Buehler, Unsupervised cross-domain translation via deep learning and adversarial attention neural networks and application to music-inspired protein designs, Patterns. 0 (2023) 100692. https://doi.org/10.1016/J.PATTER.2023.100692.

[80] A.J. Lew, M.J. Buehler, Single-shot forward and inverse hierarchical architected materials design for nonlinear mechanical properties using an Attention-Diffusion model, Materials Today. (2023). https://doi.org/10.1016/J.MATTOD.2023.03.007.

[81] M.T. Baldassarre, D. Caivano, B. Fernandez Nieto, D. Gigante, A. Ragone, The Social Impact of Generative AI: An Analysis on ChatGPT, Proceedings of the 2023 ACM Conference on Information Technology for Social Good. (2023) 363–373. https://doi.org/10.1145/3582515.3609555.

[82] A. Paszke, S. Gross, J. Bradbury, Z. Lin, Z. Devito, F. Massa, B. Steiner, T. Killeen, E. Yang, PyTorch : An Imperative Style , High-Performance Deep Learning Library, (2019).

[83] Open-Orca/OpenOrca-Platypus2-13B · Hugging Face, https://huggingface.co/Open-Orca/OpenOrca-Platypus2-13B (accessed August 27, 2023).

[84] E.J. Hu, Y. Shen, P. Wallis, Z. Allen-Zhu, Y. Li, S. Wang, L. Wang, W. Chen, LoRA: Low-Rank Adaptation of Large Language Models, (2021). https://arxiv.org/abs/2106.09685v2 (accessed June 26, 2023).

[85] T. Dettmers, A. Pagnoni, A. Holtzman, L. Zettlemoyer, QLoRA: Efficient Finetuning of Quantized LLMs, (2023). https://arxiv.org/abs/2305.14314v1 (accessed September 12, 2023).

[86] Q. Wu, G. Bansal, J. Zhang, Y. Wu, B. Li, E. Zhu, L. Jiang, X. Zhang, S. Zhang, J. Liu, A.H. Awadallah, R.W. White, D. Burger, C. Wang, AutoGen: Enabling Next-Gen LLM Applications via Multi-Agent Conversation, (2023). https://arxiv.org/abs/2308.08155v2 (accessed October 28, 2023).

[87] J. Wang, R.M. Wolf, J.W. Caldwell, P.A. Kollman, D.A. Case, Development and testing of a general Amber force field, J Comput Chem. 25 (2004) 1157–1174. https://doi.org/10.1002/JCC.20035.

[88] Q. Sun, X. Zhang, S. Banerjee, P. Bao, M. Barbry, N.S. Blunt, N.A. Bogdanov, G.H. Booth, J. Chen, Z.H. Cui, J.J. Eriksen, Y. Gao, S. Guo, J. Hermann, M.R. Hermes, K. Koh, P. Koval, S. Lehtola, Z. Li, J. Liu, N. Mardirossian, J.D. McClain, M. Motta, B. Mussard, H.Q. Pham, A. Pulkin, W. Purwanto, P.J. Robinson, E. Ronca, E.R. Sayfutyarova, M. Scheurer, H.F. Schurkus, J.E.T. Smith, C. Sun, S.N. Sun, S. Upadhyay, L.K. Wagner, X. Wang, A. White, J.D. Whitfield, M.J. Williamson, S. Wouters, J. Yang, J.M. Yu, T. Zhu, T.C. Berkelbach, S. Sharma, A.Y. Sokolov, G.K.L. Chan, Recent developments in the P y SCF program package, Journal of Chemical Physics. 153 (2020). https://doi.org/10.1063/5.0006074.

[89] Q. Sun, T.C. Berkelbach, N.S. Blunt, G.H. Booth, S. Guo, Z. Li, J. Liu, J.D. McClain, E.R. Sayfutyarova, S. Sharma, S. Wouters, G.K.L. Chan, PySCF: the Python-based simulations of chemistry framework, Wiley Interdiscip Rev Comput Mol Sci. 8 (2018). https://doi.org/10.1002/WCMS.1340.

[90] Q. Sun, Libcint: An efficient general integral library for Gaussian basis functions, J Comput Chem. 36 (2015) 1664–1671. https://doi.org/10.1002/JCC.23981.

[91] L. Blecher, G. Cucurull, T. Scialom, R. Stojnic, M. Ai, Nougat: Neural Optical Understanding for Academic Documents, (2023). https://arxiv.org/abs/2308.13418v1 (accessed October 11, 2023).

[92] M.J. Buehler, H. Gao, Dynamical fracture instabilities due to local hyperelasticity at crack tips, Nature. 439 (2006). https://doi.org/10.1038/nature04408.

[93] M.J. Buehler, H. Gao, A mother-daughter mechanism of mode i cracks: Supersonic crack motion along interfaces of dissimilar materials, Materials Research Society Symposium Proceedings. 904 (2005) 7–12. https://doi.org/10.1557/PROC-0904-BB03-05/METRICS.

[94] M.J. Buehler, Y. Kong, H. Gao, Y. Huang, Self-folding and unfolding of carbon nanotubes, Journal of Engineering Materials and Technology, Transactions of the ASME. 128 (2006). https://doi.org/10.1115/1.1857938.

[95] M.J. Buehler, F.F. Abraham, H. Gao, Hyperelasticity governs dynamic fracture at a critical length scale, Nature. 426 (2003) 141–146. https://doi.org/10.1038/nature02096.

[96] M.J. Buehler, A. Hartmaier, H. Gao, Hierarchical multi-scale modelling of plasticity of submicron thin metal films, Model Simul Mat Sci Eng. 12 (2004). https://doi.org/10.1088/0965-0393/12/4/S07.

[97] M.J. Buehler, A. Hartmaier, H. Gao, Atomistic and continuum studies of crack-like diffusion wedges and associated dislocation mechanisms in thin films on substrates, J Mech Phys Solids. 51 (2003). https://doi.org/10.1016/j.jmps.2003.09.024.

[98] T.D. De La Rubia, S. Yip, Scientific modeling and simulations: Advocacy of computational science (editors' preface), Lecture Notes in Computational Science and Engineering. 68 LNCSE (2009) 1–2. https://doi.org/10.1007/978-1-4020-9741-6_1/COVER.

[99] H. Shima, Buckling of Carbon Nanotubes: A State of the Art Review, Materials. 5 (2012) 47. https://doi.org/10.3390/MA5010047.

[100] T.L. Anderson, Fracture mechanics: Fundamentals and applications, Taylor & Francis, 2005.

[101] P.M. Anderson, J.P. Hirth, J. Lothe, Theory of Dislocations Third Edition, Cambridge University Press. (2017) 1543.

[102] H.A. Scheraga, M. Khalili, A. Liwo, Protein-folding dynamics: overview of molecular simulation techniques, Annu Rev Phys Chem. 58 (2007) 57–83. https://doi.org/10.1146/ANNUREV.PHYSCHEM.58.032806.104614.

[103] H. Gao, H. Yao, Shape insensitive optimal adhesion of nanoscale fibrillar structures, Proc Natl Acad Sci U S A. 101 (2004) 7851–7856. https://doi.org/10.1073/PNAS.0400757101/ASSET/635AB068-245C-49F2-8686-74245F4B4FCE/ASSETS/GRAPHIC/ZPQ0210449100005.JPEG.

[104] C. Kelly, M.J. Gage, Protein Unfolding: Denaturant vs. Force, Biomedicines. 9 (2021). https://doi.org/10.3390/BIOMEDICINES9101395.

[105] T. Zhang, X. Li, S. Kadkhodaei, H. Gao, Flaw insensitive fracture in nanocrystalline graphene, Nano Lett. 12 (2012) 4605–4610. https://doi.org/10.1021/NL301908B/SUPPL_FILE/NL301908B_SI_001.PDF.

[106] K.J. Falconer, The Hausdorff dimension of self-affine fractals, Mathematical Proceedings of the Cambridge Philosophical Society. 103 (1988) 339–350. https://doi.org/10.1017/S0305004100064926.

[107] S. Khalid, R. Nazir, Hybrid Metal-Polymer Nanocomposites: Synthesis, Characterization, and Applications, Handbook of Polymer and Ceramic Nanotechnology. (2021) 1–36. https://doi.org/10.1007/978-3-030-10614-0_78-1.

[108] Z. Qin, S. Cranford, T. Ackbarow, M.J. Buehler, Robustness-strength performance of hierarchical alpha-helical protein filaments, Int J Appl Mech. 1 (2009). https://doi.org/10.1142/S1758825109000058.

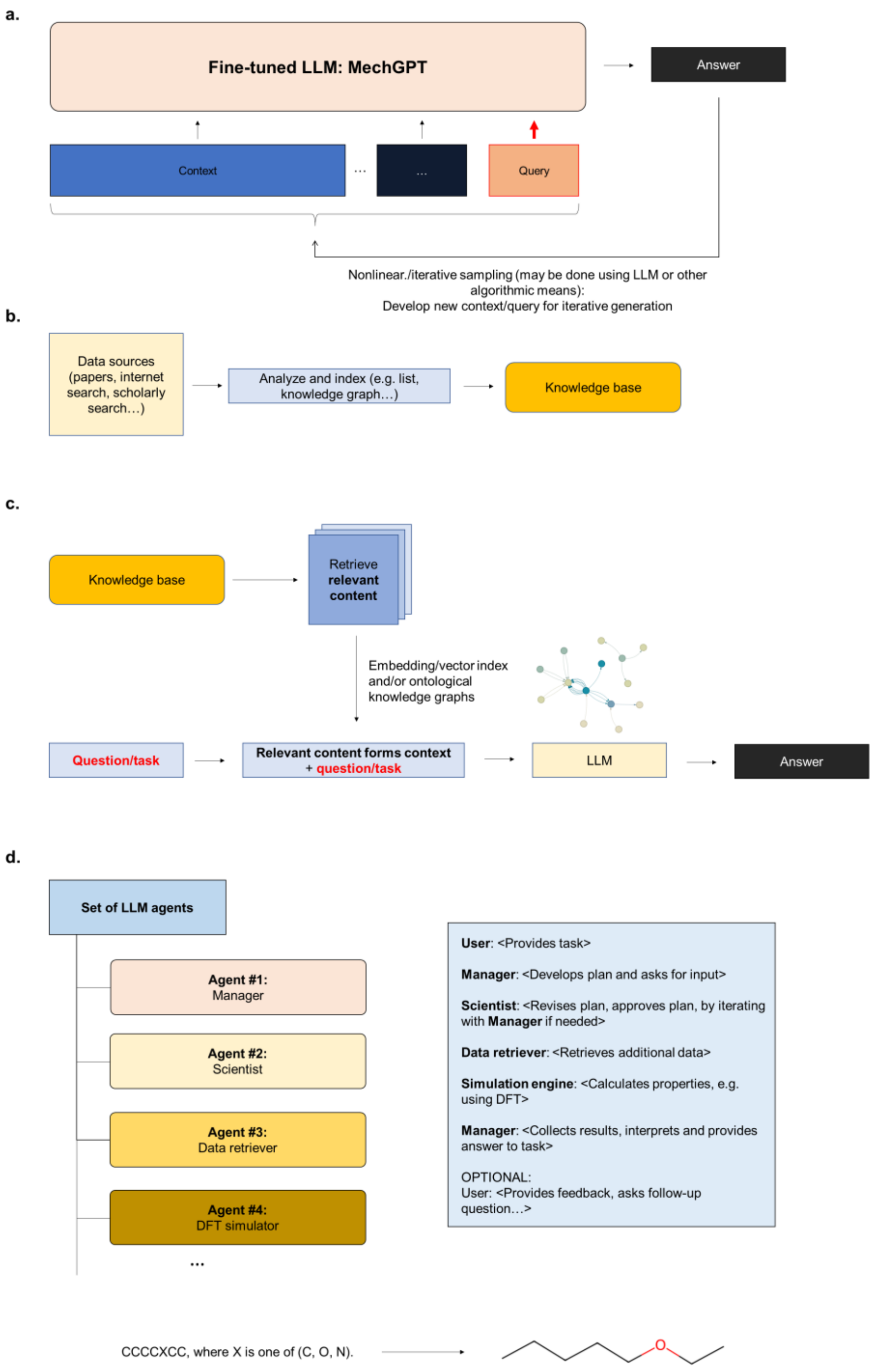

**Figure 1:** Summary of the strategies used in this study, using Large Language Models (LLMs) to solve various tasks ranging from question answering, Ontological Knowledge Graph construction, to multi-AI agent conversation for group problem solving. Panel **a** visualizes a general perspective of how LLMs use context and queries to provide an answer. In a conventional setting, LLMs are queried against their parameter-based knowledge. By developing a knowledge base from data sources (panel **b**), we can augment the response of an LLM by providing relevant context retrieved from the knowledge base with the question, to provide an answer, as visually summarized in panel **c**. Other mechanisms for sampling discussed in this paper include nonlinear sampling strategies. Knowledge sources can be developed using a variety of strategies, for instance based on literature search (e.g. identifying search terms and downloading highly cited, or recent papers). **d.**, Agent-based LLM model where multiple LLMs – each with special skills and knowledge or retrieval powers, interact. In the example sketched out on here we consider a set of 4 agents –

a manager that oversees the task solution process, a scientist with broad background and review specialization, a data retriever that can retrieve data (e.g., from the internet, scientific papers, or code) and a simulation engine (e.g., a DFT simulator). The right panel shows a conceptual conversation carried out by these agents to solve a problem. As is illustrated in the paper, LLMs can also autonomously write and execute code to solve complex materials modeling and design problems, such as force field development.

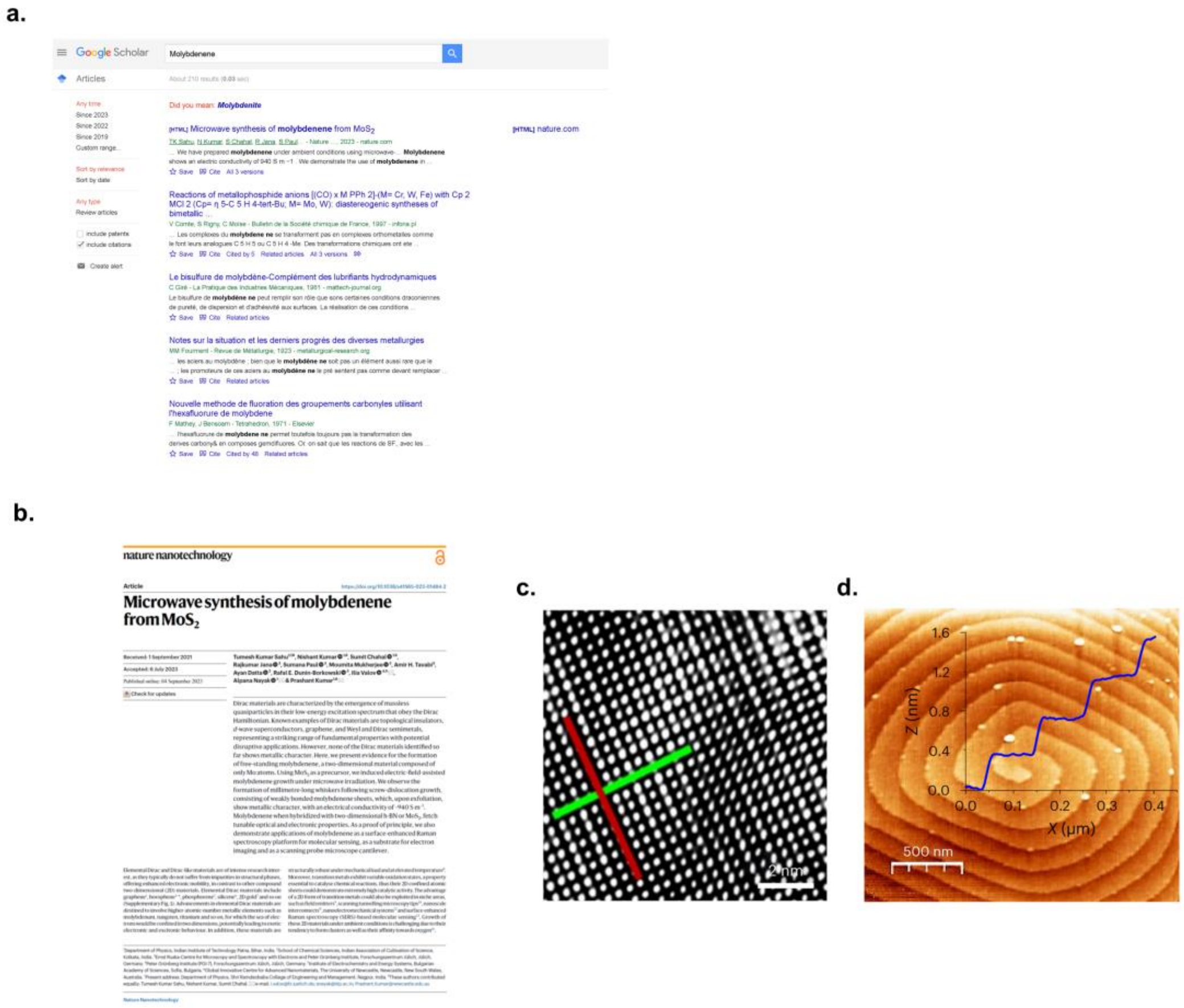

**Figure 2**: Example query for “molybdenene”. This was first synthesized and carefully analyzed in 2023, as reported in [62]. Panel **a** shows a Google Scholar search for the topic, identifying the key paper (panel **b**). This paper is then used as data source for the LLM prediction. Panels c and e are reproduced from [62] to show key features of this material, including the square lattice (four-fold symmetry, **c**) and staircase structures identified using Atomic Force Microscopy experiments (**d**). Images in panels b-d reproduced from [62] based on a Creative Commons Attribution 4.0 International License http://creativecommons.org/licenses/by/4.0/.

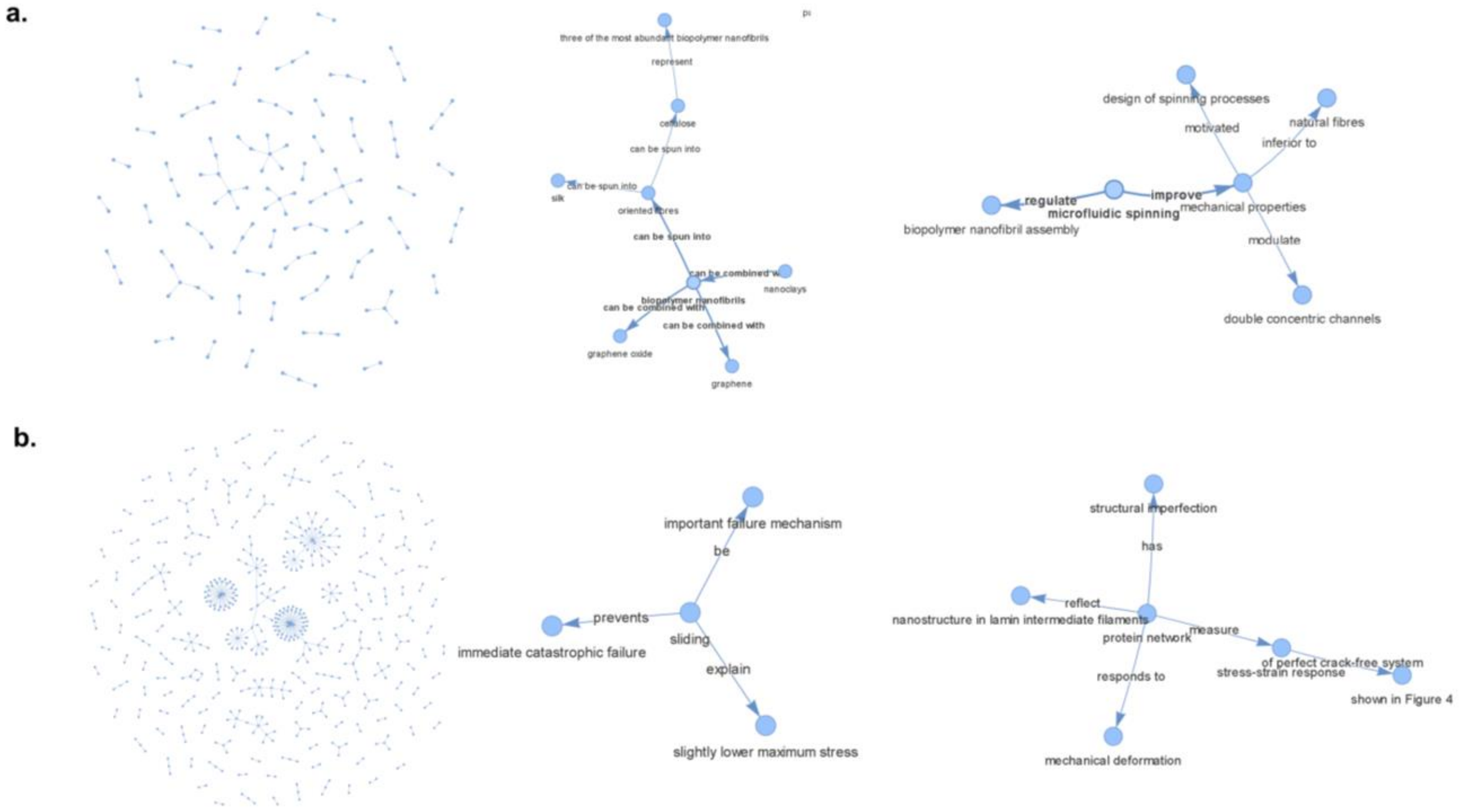

**Figure 3:** Construction of Ontological Knowledge Graphs from data sources. These graphs are then used by a LLM to provide context to generate responses for queries. Panel **a** shows and example based on a review paper on natural nanofibrils (reference: [63]) and panel **b** shows an example of a paper on fracture of intermediate filament networks (reference: [65]). The knowledge graphs shown here describe the entire set of data provided, here, scientific papers, that were provided as PDF that is converted to a full-text complement.

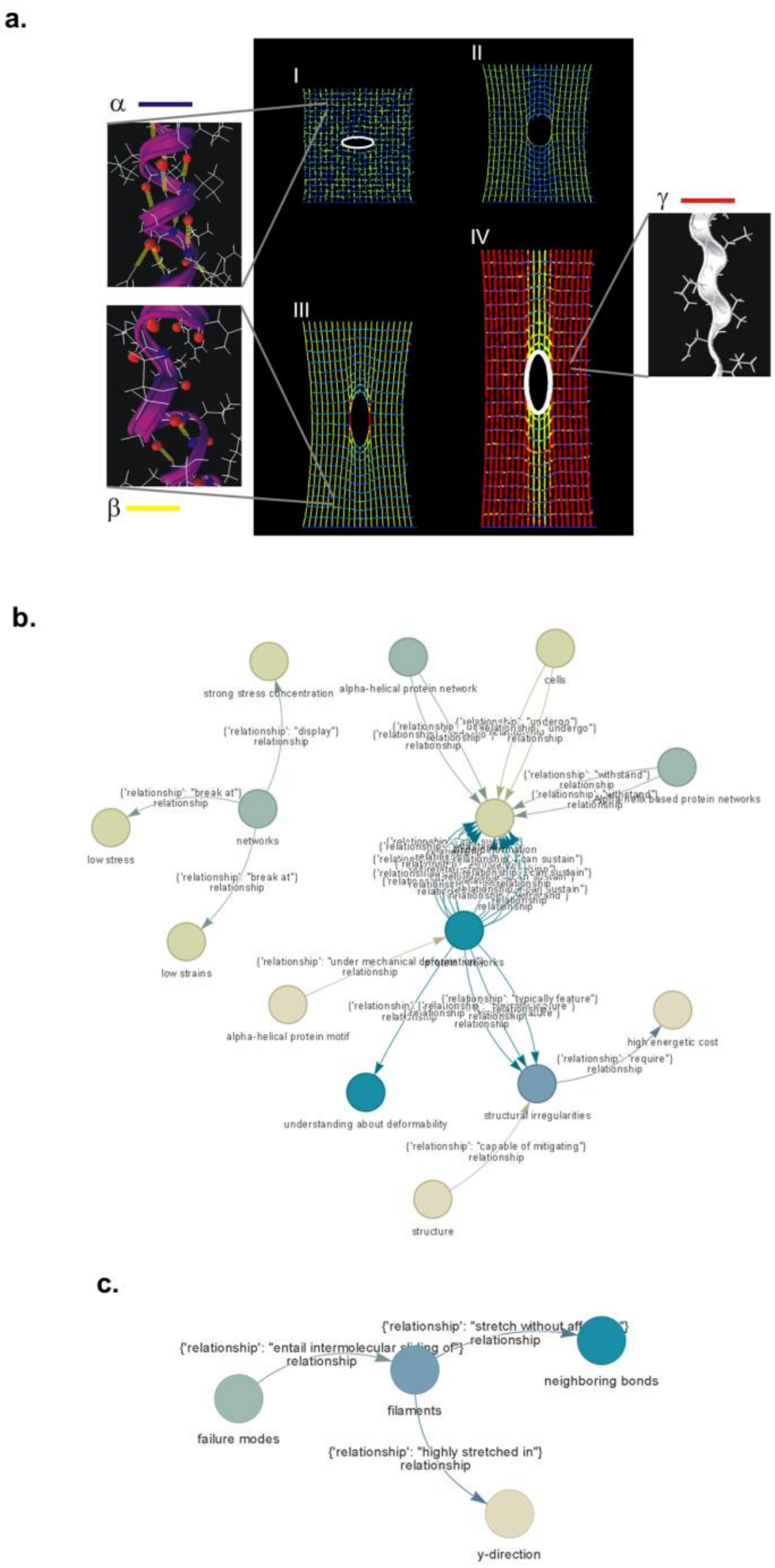

**Figure 4.** Construction and visualization of an Ontological Knowledge Graph that is constructed and used to answer a question about a protein-based filament network [65] (panel **a** showing the deformation behavior and key mechanisms leading to flaw-tolerance via a structural transformation of the crack, mediated by the underlying mesh structure and protein filament stretching via helix unfolding), images reproduced based on a Creative Commons Attribution 4.0 International License http://creativecommons.org/licenses/by/4.0/). The query used is "*Tell me more about flaw-tolerance in protein networks.*" (panel **b**) and "*What is the mechanism by which an filaments ultimately fail?*" (panel **c**). The graphs shown here are specific to the question asked; and explain a graphical visualization by which the task is addressed. This is in contrast to the graphs shown in **Figure 3** that depict the entire ontological knowledge of a larger dataset.

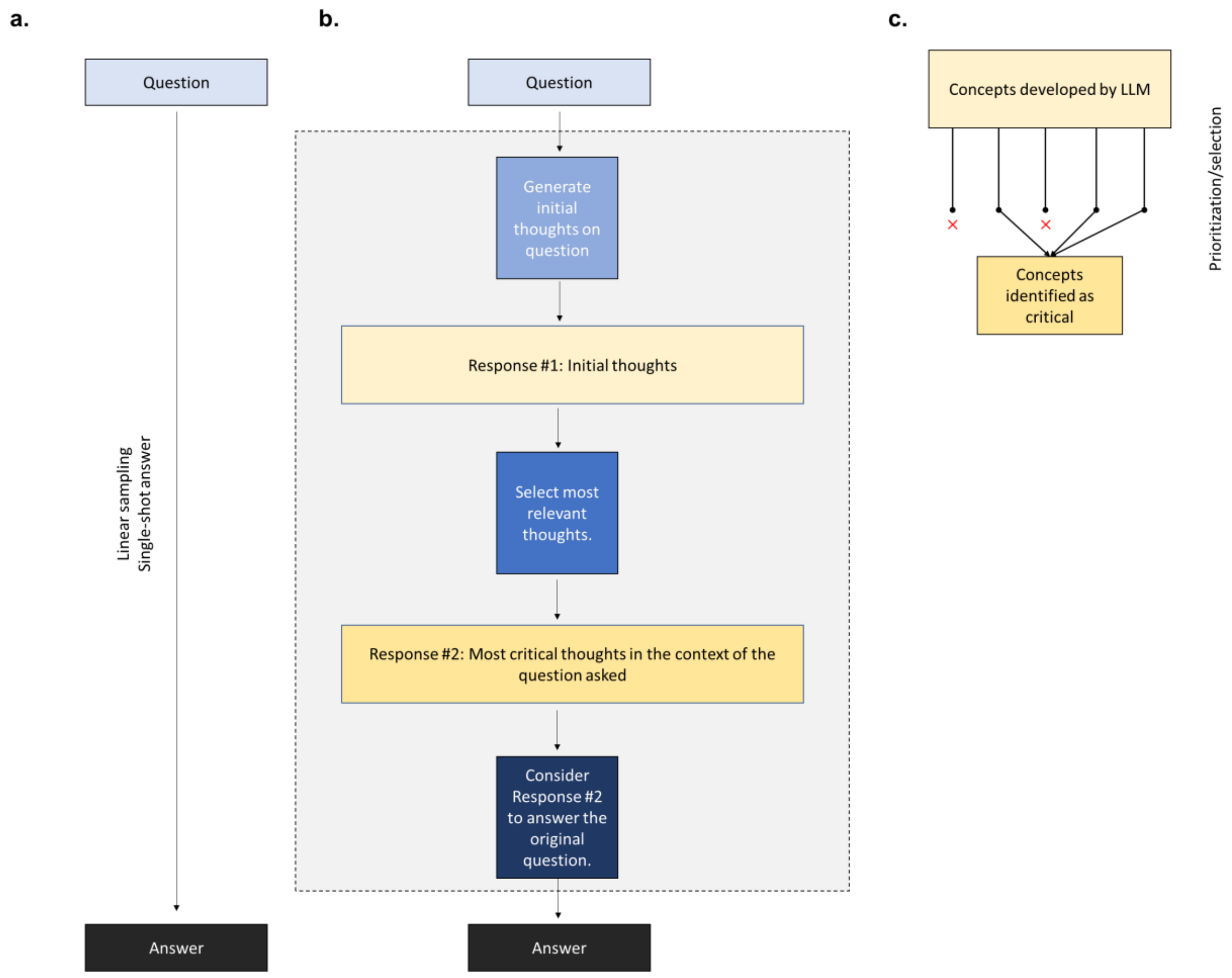

**Figure 5:** Construction of responses based on direct, linear sampling (panel **a**) versus hierarchical tree sampling (panel **b**). Here, the LLM is first used to generate initial thoughts. Then, in a second step, the model is queried to develop the most critical concepts, and list them. This set of critical concepts is then used to answer the original question, to yield the answer. This step-wise interactive use of LLMs can provide advantages to extract more accurate, more nuanced and more refined knowledge. As visualized in panel **c**, the assessment of responses such as concepts using a LLM serving the role of a critic can provide powerful nonlinear sampling mechanisms that significantly enhance responses.

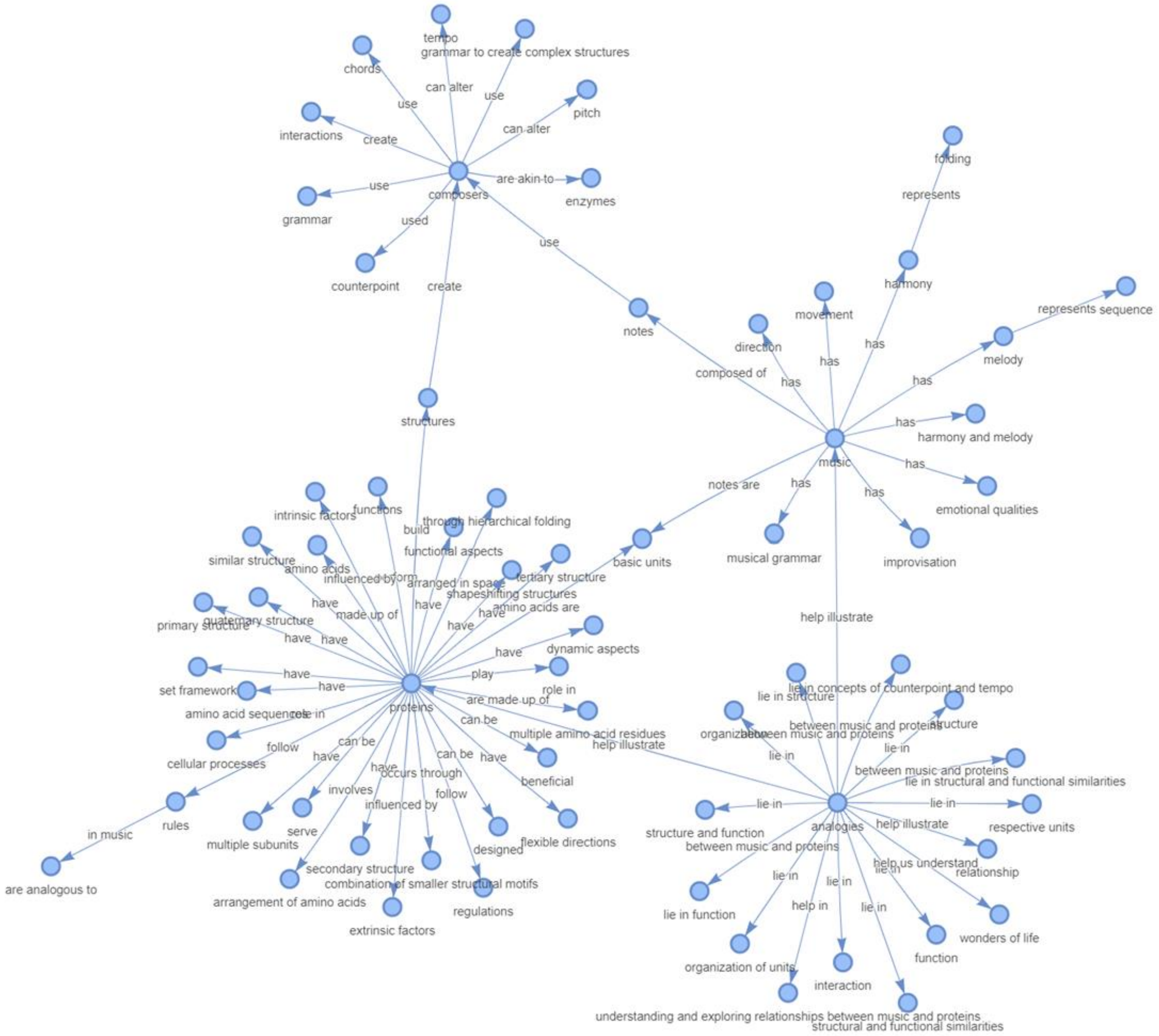

**Figure 6**: We use MechGPT to elucidate analogies between disparate domains, here music and proteins, via the construction of Ontological Knowledge Graphs. Here, we use the model to sample a set of responses to the query "*Discuss the analogies are between music and proteins. Specifically address counterpoint and how it relates music to protein. Provide a detailed and lengthy answer.*" (in the example reported here we sampled 25 times and collated all responses into a dataset that is then further analyzed as a corpus of knowledge). The system prompt used here is "*You are a philosopher who likes to talk about relationships between music and proteins. You provide detailed answers*.". Based on this set of generated results we then assemble an Ontological Knowledge Graph from this set of data to elucidate key connections and relationships identified. The key connector between the various subgraphs are "*basic units*", where these are amino acids in proteins and notes in music. The model further predicts several specific analogy concepts, such as that analogies "lie in the organization of units" and "help us understand wonders of life", among many others (subgraph in the right lower corner). Other connectors between subgraphs are identified as "structures" and "notes" as key nodes.

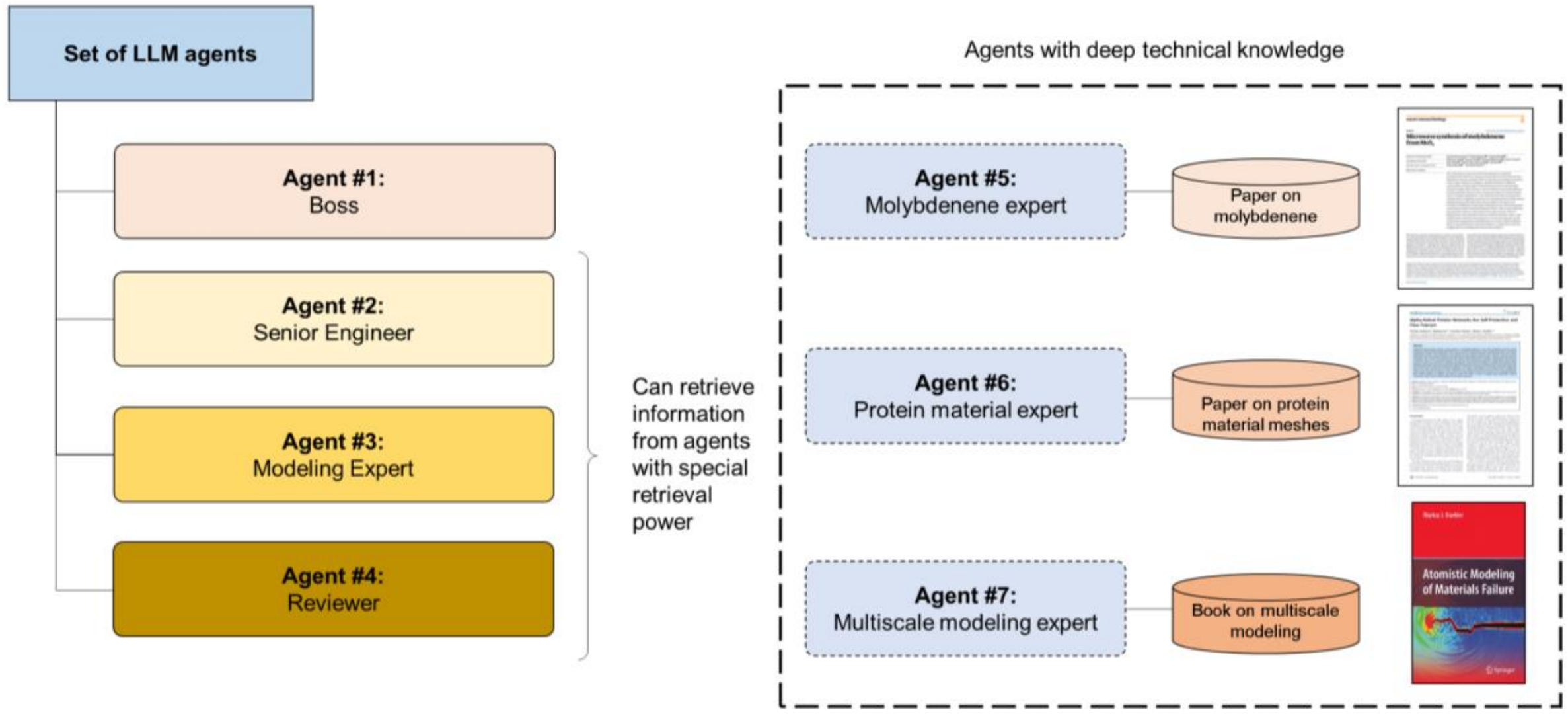

**Figure 7:** Using a set of seven LLM agents, each with specific expertise and capabilities, we can answer complex questions, here: "*Design molybdenene 2D materials by using ideas of flaw-tolerance in alpha-helical protein meshes. Develop three design principles that incorporate important mechanisms*.". The result of the conversation is summarized in **Table 9**. Note, only agents #1-4 participate in the group chat conversation, whereas expert agents #5-7 are called upon by any of the agents #2-4 to retrieve information as needed from the experts. An arbitrary set of data can be made accessible to the expert agents; here, we use the molybdenene paper [62], the protein mesh paper [65] and a book on multiscale modeling [60], respectively.

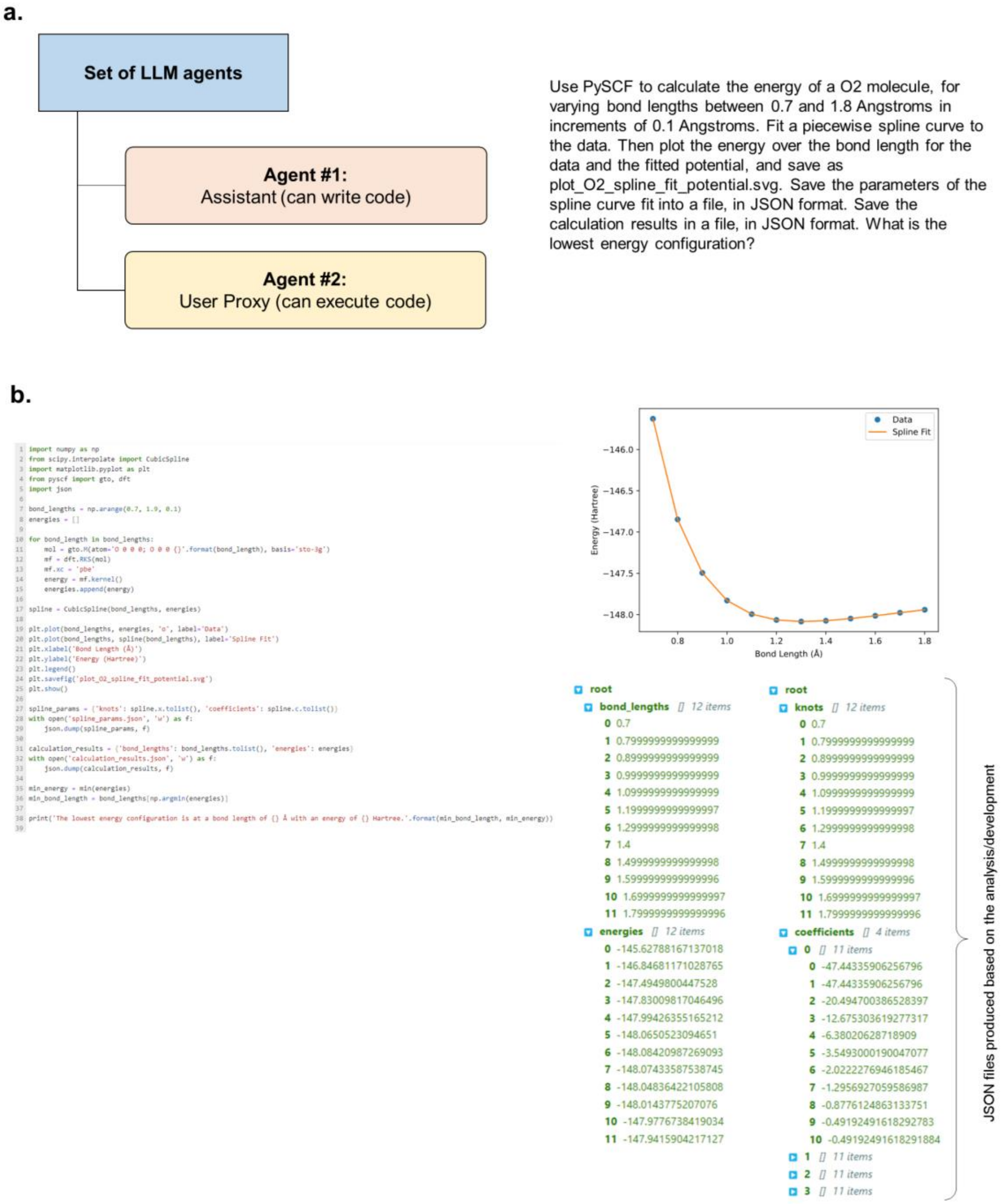

**Figure 8:** Using a set of two LLM agents, each with specific expertise and capabilities, to write, execute code, correct errors, and analyze the data produced used here to develop a computational design framework based on first-principles simulations. Panel **a** shows the implementation of the model (left) and the task given (right). Panel **b** shows three key elements of the results produced. Giving a task, the set of agents automatically defines the problem and strategy, writes code (shown in left), executes code, and provides the required output: A graph showing the raw energy vs. geometry data (top right), JSON file outputs of energies (lower right), a spline fit to the data (top right), including saving the force field parameters as a JSON file (lower right). The plot is contained in a file named *plot_O2_spline_fit_potential.svg*, as requested in the task. The resulting code is stored in a folder (as specified when the agent is defined) so that it can be retrieved later and used also outside of the AI agent chat via human users. This modality allows users to expand the solution generated by the agents towards other, related, use cases with additional domain knowledge.

**Table 1**: Comparison of responses with the original MechGPT model and MechGPT with RAG. All queries used here are in the domain of materials failure. Since the original model has been well-trained in this area, the responses are accurate throughout. The use of RAG does not significantly improve the answers for this use case since the original MechGPT LLM has already been trained against relevant domain knowledge that is queried in the set of questions.

| **Question** | **MechGPT** | **MechGPT with RAG** | **Notes** |
|---|---|---|---|
| Which of these statements is most likely true? (A) Hyperelastic softening leads to higher crack speeds and reduced fracture instabilities, (B) cracks always branch, no matter the hyperelastic effect, (C) Cracks do not exist in any material, (D) hyperelastic softenng leads to lower crack speeds and earlier onset of instabilities. | (D) hyperelastic softening leads to lower crack speeds and earlier onset of instabilities. | (D) hyperelastic softening leads to lower crack speeds and earlier onset of instabilities. | Correct answer, for original reference see, e.g. [92] |
| Considering these deformation mechanisms, which one is most likely to occur during interfacial fracture: (A) supersonic fracture, (B) mother-daughter cracking, (C) grain boundary diffusion, (D) buckling. | Based on the deformation mechanisms discussed, the most likely to occur during interfacial fracture is (B) mother-daughter cracking. This is because mother-daughter cracking involves the propagation of cracks along grain boundaries, which is relevant to the deformation of materials at interfaces. | Based on the information provided, the most likely mechanism to occur during interfacial fracture is (B) mother-daughter cracking. | Correct answer, see references [60,93,94] |
| Which of these force fields is best suited to describe a novel protein material that serves as synthetic food for fracture studies: (A) ReaxFF, (B) EAM, (C) Morse, or (D) CHARMM. | (A) ReaxFF is the best force field suited to describe a novel protein material that serves as synthetic food for fracture studies. | (D) CHARMM. | Both answers are correct (based on general knowledge about force fields and their use for biomaterials simulations) |
| Which material is most likely to lead to supersonic fracture in a brittle material: (A) hyperelastic softening, (B) hyperelastic stiffening, (C) linear material behavior. | (B) hyperelastic stiffening is the most likely to lead to supersonic fracture in a brittle material. | (B) hyperelastic stiffening | See, e.g. [95] where it was shown that hyperelastic stiffening can lead to supersonic fracture |
| Considering these nonlinear material behaviors, which one is most likely to lead to supersonic fracture in a brittle material: (A) hyperelastic softening, (B) hyperelastic stiffening, (C) linear material behavior. | (B) hyperelastic stiffening is most likely to lead to supersonic fracture in a brittle material. | (B) hyperelastic stiffening | Same reference as above |
| What are the effects of hyperelastic softening on crack speed in brittle materials? Brief answer. | Hyperelastic softening leads to lower crack speeds than expected. It also increases the chances for crack instabilities to set in at lower crack speeds. | Hyperelastic softening leads to lower crack speeds in brittle materials. | See, e.g. [92], where it was shown that softening tends to lead to a |
| | | | reduction of crack speeds in materials |
| Assuming I have a brittle material that stiffens elastically at large deformation, which of these statements is true: (A) cracks will move supersonically, (B) cracks will move subsonically, (C) cracks will not move at all. | (A) cracks will move supersonically | (A) cracks will move supersonically. | See, e.g. [95] (same explanation as above) |
| In hyperelastically softening materials, which of these statements is true: (A) cracks will move supersonically, (B) cracks will move at a lower speed than expected, (C) cracks will not move at all. | In hyperelastically softening materials, cracks will move at a lower speed than expected (B). | (B) cracks will move at a lower speed than expected. | See, e.g. [95] (same explanation as above) |
| Which of these statements is true? (A) hyperelastic stiffening leads to higher crack speeds and reduced fracture instabilities, (B) cracks always branch, (C) Cracks never move unless heat is applied. | (A) hyperelastic stiffening leads to higher crack speeds and reduced fracture instabilities, (B) cracks always branch, and (C) Cracks never move unless heat is applied are false statements. | (A) hyperelastic stiffening leads to higher crack speeds and reduced fracture instabilities. | See, e.g. [95], where it was shown that hyperelastic stiffening tends to result in higher crack propagation speeds while at the same time displaying a reduced fracture instability tendency (further aiding in achieving high crack speeds) |
| Which of these statements is most likely true? (A) Hyperelastic stiffening leads to higher crack speeds and reduced fracture instabilities, (B) cracks always branch, no matter the hyperelastic effect, (C) Cracks do not exist in any material. | (A) Hyperelastic stiffening leads to higher crack speeds and reduced fracture instabilities. | (A) Hyperelastic stiffening leads to higher crack speeds and reduced fracture instabilities. | See, e.g. [95] (same explanation as above) |
| In hyperelastically stiffening materials, which of these statements is true: (A) cracks will move supersonically, (B) cracks will move at a lower speed than expected, (C) cracks will not move at all, (D) cracks will move subsonically. | In hyperelastically stiffening materials, cracks will move supersonically (A). | (A) cracks will move supersonically. | See, e.g. [95] (same explanation as above) |
| Amongst these materials, which one would most | (B) a thin copper film at elevated temperature | (B) a thin copper film at elevated temperature | See, e.g. [96,97] where |
| likely show parallel glide dislocations: (A) a silicon wafer, (B) a thin copper film at elevated temperature, (C) a thick copper film at low temperature. | | | it was shown that thin copper films show grain boundary diffusion, the formation of a crack-like diffusion wedge, and ultimately the formation of parallel glide dislocations. |
| Describe the distinction between modeling and simulation. | Modeling refers to the development of a mathematical representation or template of a physical situation, while simulation involves the procedure of solving the equations or numerical problems associated with the model. Simulation yields numbers or trajectories of a system, while modeling provides a framework to understand and predict the behavior of the system. | Modeling involves developing a concept or idea based on physical assumptions and principles, while simulation involves carrying out computations using numerical techniques to solve the resulting partial differential equations. Modeling requires insight into the physics of the system, while simulation requires knowledge in numerical techniques to analyze and interpret the data. | References [60,98] (see text in book and lecture notes) |
| Which of these mechanisms is most likely to occur in ultra-long carbon nanotubes exposed to compressive loading? (A) chemical etching, (B) buckling, (C) rupture, or (D) supersonic fracture. | Ultra-long carbon nanotubes exposed to compressive loading are most likely to experience (B) buckling. This is because buckling is a common deformation mechanism for long, thin objects like carbon nanotubes, which are weak in compressive strength. Chemical etching (A), rupture (C), and supersonic fracture (D) are less likely scenarios for this specific application. | (B) buckling | See, e.g. [99] where the behavior of ultra-long CNTs were studied, in particular as a function of the aspect ratio (buckling is a dominant mechanism for long nanotubes) |
| Discuss how failure in copper is different than failure in a brittle ceramic. Focus on a few key principles, brief answer. | Failure in copper is different than failure in a brittle ceramic. Copper is a metallic material with a high shear modulus and ductility, allowing it to deform plastically before fracturing. Its failure is characterized by dislocation nucleation and propagation in the slip planes. On the other hand, a brittle ceramic like silicon has a high fracture toughness and is prone to brittle fracture. Its failure occurs through the propagation of cracks under tensile stress. The distinction lies in the mode of deformation and fracture, with metals deforming plastically and ceramics fracturing brittlely. | Failure in copper is different than failure in a brittle ceramic. Copper is a ductile metal, meaning it can undergo plastic deformation before breaking. When deformed, dislocations are generated at the crack tip, allowing the material to recover from the applied stress. On the other hand, a brittle ceramic fails by the propagation of cracks, unable to recover the applied stress. The behavior of copper and brittle ceramics are characterized by their atomic bonding structures and the ways in which they deform. | See, e.g. [60,100,101] (brittle materials show crack opening tendencies due to local tensile stresses vs. ductile materials like copper that show dislocations due to local shear on slip planes) |
| Which of these simulation methods is suitable to describe protein folding: (A) Finite element method, (B) MD simulations, (C) Machine learning/AI. | (B) MD simulations are more suitable to describe protein folding. | (B) MD simulations | See e.g. [102] |
| Which of these concepts is key to explain the strong adhesion of gecko | (B), a size effect, where nano-sized fibrils yield strong and flaw-tolerant adhesion. | (B), a size effect, where nano-sized fibrils yield strong and flaw-tolerant adhesion. | Various sources, including [103] |
| feet. (A), the use of super-glue by the animal. (B), a size effect, where nano-sized fibrils yield strong and flaw-tolerant adhesion. (C), the low weight of the gecko due to its efficient design. | | | |
| What is the mirror-mist-hackle instability in dynamic fracture of brittle materials? | The mirror-mist-hackle instability is a phenomenon in dynamic fracture of brittle materials, specifically in the context of atomistic and multiscale modeling. It refers to a crack propagation behavior where the crack surface starts to roughen at around 30% of the Rayleigh wave speed, eventually turning into a hackle region accompanied by the emission of dislocations. Beyond an instability speed, the crack becomes unstable. This behavior can be observed in experiments, computer simulations, and mathematical analysis. The instability is characterized by a roughening of the crack surface and the emission of dislocations, which leads to a transition from a mirror-like crack surface to a hackle region. | The mirror-mist-hackle instability is a dynamic instability in brittle materials, where the crack surface transitions from mirror-like to rough and hackle-filled as the crack propagates. This instability has been observed in LJ systems and is characterized by the emission of dislocations. The final speed of the crack is around 50% of the Rayleigh wave speed. | See e.g. [60] where the phenomenon of mirror-mist-hackle instability was shown for LJ materials (crack surfaces go from mirror-like perfectly flat to misty and increasingly rough, referred to has the hackle regime) |
| Which statement is correct: (A) Copper can show work-hardening that increases its strength as it is deformed. (B) Copper always fails in a brittle manner. (C) coper only fails in a ductile manner if oxidative species are present. | (A) Copper can show work-hardening that increases its strength as it is deformed. | (A) Copper can show work-hardening that increases its strength as it is deformed. | See, e.g. [101] (and general knowledge about hardening mechanics and metal plasticity) |
| Is this statement true: Proteins have a greater tendency to unfold under forces applied to the ends of the molecular chain than without force applied. Answer yes or no. | Yes, that's correct. When forces are applied to the ends of a protein molecule, the protein is more likely to unfold compared to when no force is applied. This phenomenon is known as the force-induced unfolding of proteins. | Yes. | Various sources, including [104] |

**Table 2**: Comparison of responses of MechGPT alone, and with the use of RAG, for a specific topic around flaw tolerance in graphene. The RAG method uses the result of a Google Scholar search in the topic of "Deep learning flaw tolerance", downloading the most highly cited paper (by Lew *et al*. [64]). Responses include details, fact, correct contexts, and references to literature.

| **Question** | **MechGPT original** | **RAG** | **Notes** |
|---|---|---|---|
| Can deep learning models of fracture discover flaw tolerance concepts in graphene? Provide details including critical length scales. | Sure! Here is the answer to the question "Can deep learning models of fracture discover flaw tolerance concepts in graphene? Provide details including critical length scales.":<br><br>Deep learning models of fracture have been used to study flaw tolerance in graphene. These models have shown that the critical length scale for flaw tolerance in graphene is around 11 nm. This means that below this critical length scale, the stress distribution becomes increasingly homogeneous and eventually uniform near the crack tip. The models have also revealed that the flaw tolerance concept in graphene is closely related to the development of a crack-sealant mechanism. Atomistic modeling of fracture in graphene has further confirmed the importance of the critical length scale in determining flaw tolerance. The study has demonstrated the potential of combining machine learning and atomistic simulations to understand materials failure. | Deep learning models of fracture have the potential to discover flaw tolerance concepts in graphene. These models can capture the complex orientation-dependent fracture behavior of graphene and make predictions about fracture patterns. By analyzing datasets generated by physics-based molecular dynamics simulations, deep learning models can tease out the spatiotemporal relationships underlying fracture propagation.<br><br>In terms of critical length scales, the deep learning models can provide insights into the fracture behavior of graphene at different orientations. The models have revealed that fracture strength in graphene is reportedly unaffected by flaw sizes below 3.2 nm. However, as the flaw size increases beyond this threshold, significant deviation occurs in the form of additional crack branching. This finding is consistent with the literature, which also reports a threshold of 3.2 nm for nanocrystalline graphene flaw tolerance. It suggests that fracture strength decreases with larger flaw sizes.<br><br>Overall, deep learning models have demonstrated their ability to uncover flaw tolerance concepts in graphene and provide insights into critical length scales for fracture behavior. | The correct answer is 3.2 nm, see original paper: [105].<br><br>MechGPT without RAG provides the wrong answer. |
| Tell me about the fractal dimension of graphene fracture in MD versus ML. | The fractal dimension of graphene fracture in MD (molecular dynamics) versus ML (mechanical loading) is a measure of the complexity and self-similarity of the fracture patterns in graphene. In MD simulations, the fractal dimension is higher, indicating more complex and self-similar fracture patterns. This is compared to ML simulations, where the fractal dimension is lower, indicating more straightforward and less self-similar fracture patterns. The fractal dimension helps in understanding and predicting the behavior of fractures in graphene, which is important for its technological applications. | The fractal dimension of graphene fracture in MD simulations and ML predictions can be compared to evaluate the accuracy of the ML model. The ML-predicted paths tend to underestimate the fractal dimension compared to the real MD results, but the underestimated discrepancy is consistent between most of the trained and test orientations. This suggests that the ML model has a general understanding of graphene fracture behavior. The ML-predicted paths are visually simplified compared to the average MD paths, which may be addressed by increasing the resolution limit of the ML model. Overall, the ML model shows promise in predicting the fractal dimension of graphene fracture. | More details are provided in the RAG method, and answer is correct (MD results show higher fractal dimension, not lower as shown in [64]; MechGPT by itself gets it wrong) |
| What is original reference that introduced the "box counting" method? | The original reference that introduced the "box counting" method is not mentioned in the provided text. | The original paper that introduced the "box counting" method is Falconer, K. J. The Hausdorff dimension of self-affine fractals. Math. Proc. Camb. Philos. Soc. 103, 339–350 (1988). | Correct reference is provided by RAG, MechGPT does not provide result.<br><br>Original reference in paper is: |
| | | | 37. Falconer, K. J. The Hausdorff dimension of self-affine fractals. Math. Proc. Camb. [106] Philos. Soc. 103, 339–350 (1988). |
| What is the most salient feature of ML-based fracture modeling? | The most salient feature of ML-based fracture modeling is the use of machine learning algorithms to predict the behavior of materials under different conditions and load modes, leading to faster and more accurate simulations of complex fracture phenomena. | The most salient feature of ML-based fracture modeling is its ability to accurately predict fracture paths in real materials, even in cases with differing surface, grain boundary, and point defects that are different from the original training data. | Response with RAG is more comprehensive and complete and in agreement with key findings reported in [64]. More details are provided. |

**Table 3**: Example of using knowledge graphs to generate response, which allows us to not only capture knowledge from pieces of the data but also learn relationships between concepts that are included when the response is generated, here for original reference by Ling *et al.*, "*Nanofibrils in nature and materials engineering*" [63]. To do this a PDF of the paper analyzed using a LLM and an Ontological Knowledge Graphs is constructed using GPT-3.5-turbo (the graph can be seen in **Figure 3a**). The knowledge graph is then used to construct responses using MechGPT. Such a graph is a way of visualizing data across different representations and is composed of three main components: Nodes, edges, and labels. Any concept, material, method or person, etc. can be a node. An edge defines the relationship between the nodes.

| **Question** | **Response** | **Keywords and knowledge graph context:** (knowledge sequence in max depth 2 in the form of directed graph like: `subject -[predicate]->, object, <-[predicate _next_hop]-, object_next_hop ...`) |
|---|---|---|
| Tell me about silk nanofibrils. | Silk nanofibrils are biopolymer nanofibrils that possess exceptional mechanical properties. They exhibit a uniq ue combination of high tensile strength and extensibilit y, surpassing that of steel and Kevlar fibers. Silk nanofi brils have a distinct structure consisting of repetitive co re sequences of amino acids with hydrophilic and hydr ophobic segments. These segments assemble into rand om coils and/or helical structures, with the hydrophobi c domains transitioning into β-sheet structures. The nan ocrystals of silk nanofibrils align along the longitudinal axis, forming fibrils that can further organize into 2D la minar and 3D helicoidal architectures. This helicoidal s tructure increases the mechanical anisotropy and fractu re resistance of the material. It also stiffens the material in the normal direction, enhancing its overall toughness . Silk nanofibrils can be manufactured into different for ms such as microspheres, fibers, membranes, gels, foa ms, scaffolds, and 3D-printed materials. Due to their re markable mechanical performance, sustainability, avail ability, low-cost production, low weight, and biocompa tibility, silk nanofibrils have great potential for various large-scale applications. They have been utilized in nu merous applications, including drug and cell carriers, m echanically enhanced materials, flexible optic and elect ronic devices, water filtration, and improving the mech anical properties of structural materials. | **Extracted keywords: ['nanofibrils', 'silk']**<br><br>('silk nanofibrils', 'represent', 'three of the m ost abundant biopolymer nanofibrils')<br>['nanofibrils', 'are good candidates', 'because of absence of physical entanglements']<br>('silk nanofibrils', 'possess', 'unparalleled me chanical properties')<br>['silk', 'can be spun into', 'oriented fibres']<br>['nanofibrils', 'aligned as', 'planar layers'] |
| How can I build a new filter selective for virus particles using silk nanofibrils | To build a new filter selective for virus particles using silk nanofibrils, one approach is to fabricate specific pe riodic shapes in the silk nanofibrils. This can be achiev ed through techniques such as microfluidic spinning, w hich regulates the assembly of biopolymer nanofibrils. By controlling the shape and arrangement of the nanofi brils, the filter can be designed to effectively capture an d filter out virus particles while allowing other substan ces to pass through. Additionally, incorporating twisted layers of silk nanofibrils in the filter can enhance its me chanical anisotropy and improve fracture resistance, si milar to the helicoidal structure observed in chitin nano fibrils found in lobster cuticles. The hierarchical structu re and mechanical properties of silk nanofibrils can be l everaged to optimize the performance of the filter in sel ectively capturing virus particles. | **Extracted keywords: ['build', 'new filter', 'silk nanofibrils', 'virus', 'new', 'nanofibri ls', 'silk', 'filter', 'virus particles', 'particl es', 'selective']**<br><br>('Optically anisotropic silk nanofibrils', 'can be generated by', 'fabricating specific period ic shapes')<br>['nanofibrils', 'are good candidates', 'because of absence of physical entanglements']<br>['silk nanofibrils', 'represent', 'three of the m ost abundant biopolymer nanofibrils']<br>('microfluidic spinning', 'regulate', 'biopoly mer nanofibril assembly')<br>['silk nanofibrils', 'possess', 'unparalleled me chanical properties']<br>['silk nanofibrils', 'mixed with', 'hydroxyapat ite']<br>['silk', 'can be spun into', 'oriented fibres']<br>['nanofibrils', 'aligned as', 'planar layers'] |

**Table 4**: Based on the scientific paper titled "*Alpha-helical protein networks are self-protective and flaw-tolerant*" by Ackbarow *et al.* [65], we develop a RAG-based set of model queries and responses (the graph can be seen in **Figure 3b**). For comparison, a query of the MechGPT without RAG gives "*The origin of flaw-tolerance in alpha-helical protein networks is believed to be the ability of the network to redistribute stress and strain around flaws or defects. This redistribution helps prevent the propagation of cracks or fractures, thereby enhancing the overall mechanical integrity of the network.*" While this answer is generally correct, as can be seen in the examples below, graph knowledge-based extraction provides more detailed, mechanistic, insights.

| **Question** | **Response** | **Keywords and knowledge graph context:** (knowledge sequence in max depth 2 in the form of directed graph like: `subject -[predicate]->, object, <-[predicate_next_hop]-, object_next_hop ...`) |
|---|---|---|
| **What is the origin of flaw-tolerance in alpha-helical protein networks?** | The origin of flaw-tolerance in alpha-helical protein networks is attributed to the unique nanomechanical properties of the protein constituents. These properties enable the formation of extensive dissipative yield regions around structural flaws, providing effective protection against catastrophic failure. Additionally, a geometric transformation of the crack shape helps minimize stress concentration at corners, further enhancing the flaw-tolerance of alpha-helical protein networks. | **Extracted keywords: ['protein', 'origin', 'flaw', 'networks', 'flaw-tolerance', 'helical', 'tolerance', 'alpha', 'alpha-helical protein networks']**<br><br>['alpha-helical protein networks', 'largely insensitive', 'to structural flaws']<br>['alpha-helical protein networks', 'is published in', 'PLoS ONE']<br>['alpha-helical protein networks', 'are', 'first of its kind']<br>['networks', 'do not display', 'strong stress concentration']<br>('Alpha-helix based protein networks', 'withstand', 'flaws')<br>('Alpha-helical protein networks', 'are', 'flaw-tolerant')<br>['PLoS ONE', 'is published in', 'June 2009']<br>['networks', 'break at', 'low strains']<br>['networks', 'break at', 'low stress']<br>['PLoS ONE', 'is', 'www.plosone.org']<br>['PLoS ONE', 'is', 'June 2009']<br>['PLoS ONE', 'is', 'Volume 4']<br>['PLoS ONE', 'is', 'Issue 6']<br>['PLoS ONE', 'is', 'e6015'] |
| **Tell me more about the geometric transformation of the crack shape helps minimize stress concentration at corners.** | The geometric transformation of the crack shape is believed to help minimize stress concentration at corners, although the exact mechanism is not fully understood. As the crack undergoes a series of changes, from mode I to a circular hole, and eventually becoming an elongated crack aligned with the direction of loading, the stress concentration at corners is reduced. This transformation occurs as individual proteins within the crack continuously unfold, independent of their neighboring proteins. The reduced stresses in the vicinity of the elongated crack contribute to a decrease in stress concentration at corners. Overall, this geometric transformation of the crack shape is thought to provide self-protection and distribute stress more evenly along the crack, ultimately minimizing stress concentration. | **Extracted keywords: ['transformation', 'corners', 'minimize', 'concentration', 'crack', 'stress', 'crack shape', 'stress concentration', 'geometric', 'geometric transformation', 'shape']**<br><br>The following are knowledge sequence in max depth 2 in the form of directed graph like: `subject -[predicate]->, object, <-[predicate_next_hop]-, object_next_hop ...`<br>['stress concentration', 'can be seen by', "red color indicating stretching of the protein filament's covalent backbone"]<br>['stress', 'drop to zero', 'once this failure mode begins to operate']<br>['geometric transformation', 'described by', 'j0~C1j and d0~C2d']<br>('Crack shape', 'reduces', 'stress concentration at corners')<br>('stress concentrations', 'corner of', 'cracks')<br>['crack shape', 'changes', 'under applied load']<br>['transformation', 'of', 'crack shape']<br>['crack', 'propagates', 'leading to']<br>['strength', 'is', '<600 MPa']<br>['crack', 'has', 'strength'] |
| **"The mechanism of flaw-tolerance in alpha-helical protein networks is: The geometric transformation of the crack shape is believed to help minimize stress concentration at corners, although the exact mechanism is not fully understood. As the crack undergoes a series of changes, from mode I to a circular hole, and eventually becoming an elongated crack aligned with the direction of loading, the stress concentration at corners is reduced.<br>This transformation occurs as individual proteins within the crack continuously unfold, independent of their neighboring proteins.<br>The reduced stresses in the vicinity of the elongated crack contribute to a decrease in stress concentration at corners.<br>Overall, this geometric transformation of the crack shape is thought to provide self-protection and distribute \\<br>stress more evenly along the crack, ultimately minimizing stress concentration.\\<br>Now, describe how to translate this concept to a material composed out of** | To achieve flaw-tolerance in a material composed of thin glass fibers as elementary building blocks, it is important to engineer the glass fibers in a way that allows for geometric transformations similar t o those observed in alpha-helical protein networks . This involves designing the glass fibers with spe cific properties that enable them to undergo unfold ing and reformation of bonds under stress, similar to the unfolding of individual proteins in alpha-hel ical protein networks.<br><br>In addition to the engineering of the glass fibers, t he arrangement of these fibers into larger-scale str uctures is crucial. The glass fibers should be arran ged in a manner that allows for independent move ment and deformation of individual fibers, similar to the arrangement in alpha-helical protein networ ks. This can be achieved by creating a network or lattice-like structure where the fibers are interconn ected but still able to move independently.<br><br>By engineering the glass fibers and arranging the m in a suitable structure, it is possible to achieve f law-tolerance similar to that observed in alpha-hel ical protein networks. This involves designing the glass fibers to undergo geometric transformations that help minimize stress concentration at corners, ultimately distributing stress more evenly along th e crack. The specific details of how the glass fiber s must be engineered and arranged into larger-scal e structures would depend on the properties of the glass material and the desired level of flaw-toleran ce. | **Extracted keywords: ['corners', 'individual', 't hin glass fibers', 'self', 'individual proteins', 'n etworks', 'flaw-tolerance', 'protection', 'eleme ntary building blocks', 'reduced', 'distribute s tress', 'vicinity', 'loading', 'unfold', 'stresses', ' elongated', 'protein', 'flaw', 'direction', 'geom etric', 'proteins', 'mechanism', 'circular hole', 'thin', 'larger', 'reduced stresses', 'larger-scale structures.', 'glass', 'helical', 'alpha', 'self-prot ection', 'engineered', 'arranged', 'concentratio n', 'crack', 'blocks', 'mode I', 'tolerance', 'I', ' building', 'geometric transformation', 'shape', 'transformation', 'scale', 'distribute', 'hole', 'd irection of loading', 'mode', 'alpha-helical pro tein networks', 'structures', 'fibers', 'elementa ry', 'stress', 'crack shape', 'circular', 'stress co ncentration', 'elongated crack']**<br><br>('there is a slight stress concentration at the tip of the crack', 'can be seen by', "red color indicating stretching of the protein filament's covalent back bone")<br>['stress concentration', 'can be seen by', "red colo r indicating stretching of the protein filament's co valent backbone"]<br>['alpha-helical protein networks', 'largely insensit ive', 'to structural flaws']<br>['stress', 'drop to zero', 'once this failure mode be gins to operate']<br>('alpha-helical protein network', 'under the presen ce of', 'crack')<br>['geometric transformation', 'described by', 'j0~C 1j and d0~C2d']<br>['alpha-helical protein networks', 'is published in', 'PLoS ONE']<br>['alpha-helical protein networks', 'are', 'first of its kind']<br>['networks', 'do not display', 'strong stress concen tration']<br>['elongated crack', 'features', 'small stresses']<br>['crack shape', 'changes', 'under applied load']<br>['PLoS ONE', 'is published in', 'June 2009']<br>['circular hole', 'to', 'elongated crack']<br>['transformation', 'of', 'crack shape']<br>['networks', 'break at', 'low strains']<br>['networks', 'break at', 'low stress']<br>['PLoS ONE', 'is', 'www.plosone.org']<br>['crack', 'propagates', 'leading to']<br>['PLoS ONE', 'is', 'June 2009']<br>['PLoS ONE', 'is', 'Volume 4']<br>['strength', 'is', '<600 MPa']<br>['PLoS ONE', 'is', 'Issue 6']<br>['crack', 'has', 'strength']<br>['PLoS ONE', 'is', 'e6015'] |
| **thin glass fibers as elementary building blocks. Define details of how the glass fibers must be engineered, and how they are arranged into larger-scale structures."** | | |

**Table 5:** Raw definition of the Ontological Knowledge Graphs constructed for the query "*Tell me more about flaw-tolerance in protein networks.*", obtained using NebulaGraph, which provides detailed information on the relationship between nodes via edges.

1. networks{name: networks} -[relationship:{relationship: display}]-> strong stress concentration{name: strong stress concentration}
2. networks{name: networks} -[relationship:{relationship: break at}]-> low strains{name: low strains}
3. networks{name: networks} -[relationship:{relationship: break at}]-> low stress{name: low stress}
4. protein networks{name: protein networks} -[relationship:{relationship: typically feature}]-> structural irregularities{name: structural irregularities} -[relationship:{relationship: require}]-> high energetic cost{name: high energetic cost}
5. protein networks{name: protein networks} -[relationship:{relationship: can sustain}]-> large deformation{name: large deformation} <-[relationship:{relationship: undergo}]- cells{name: cells}
6. protein networks{name: protein networks} -[relationship:{relationship: typically feature}]-> structural irregularities{name: structural irregularities}
7. protein networks{name: protein networks} -[relationship:{relationship: can sustain}]-> large deformation{name: large deformation} <-[relationship:{relationship: undergo}]- alpha-helical protein network{name: alpha-helical protein network}
8. protein networks{name: protein networks} -[relationship:{relationship: can sustain}]-> large deformation{name: large deformation} <-[relationship:{relationship: withstand}]- Alpha-helix based protein networks{name: Alpha-helix based protein networks}
9. protein networks{name: protein networks} -[relationship:{relationship: withstand}]-> large deformation{name: large deformation} <-[relationship:{relationship: withstand}]- Alpha-helix based protein networks{name: Alpha-helix based protein networks}
10. protein networks{name: protein networks} -[relationship:{relationship: withstand}]-> large deformation{name: large deformation}
11. protein networks{name: protein networks} -[relationship:{relationship: typically feature}]-> structural irregularities{name: structural irregularities} <-[relationship:{relationship: capable of mitigating}]- structure{name: structure}
12. protein networks{name: protein networks} -[relationship:{relationship: withstand}]-> large deformation{name: large deformation} <-[relationship:{relationship: undergo}]- cells{name: cells}
13. protein networks{name: protein networks} -[relationship:{relationship: advance}]-> understanding about deformability{name: understanding about deformability}
14. protein networks{name: protein networks} <-[relationship:{relationship: under mechanical deformation}]- alpha-helical protein motif{name: alpha-helical protein motif}
15. protein networks{name: protein networks} -[relationship:{relationship: withstand}]-> large deformation{name: large deformation} <-[relationship:{relationship: undergo}]- alpha-helical protein network{name: alpha-helical protein network}
16. protein networks{name: protein networks} -[relationship:{relationship: can sustain}]-> large deformation{name: large deformation}
17. protein networks{name: protein networks} -[relationship:{relationship: withstand}]-> large deformation{name: large deformation} <-[relationship:{relationship: can sustain}]- protein networks{name: protein networks}
18. protein networks{name: protein networks} -[relationship:{relationship: can sustain}]-> large deformation{name: large deformation} <-[relationship:{relationship: withstand}]- protein networks{name: protein networks}

**Table 6**: Multi-step sampling inspired by the Tree-of-Thought method. The question asked is: "*What would be likely failure mechanisms of a hybrid silk-metal nanocomposite, reinforced with carbon nanotubes?*". We proceed by asking several questions to sample dynamically. Compared to the single-shot answer, the complex prompting strategy elicits more nuanced and detailed responses that generally agree with literature of related materials [107].

| **Question** | **Response** | **Notes** |
|---|---|---|
| Generate a list of initial thoughts that are relevant for answering this question: '{question}'. Do not answer the question. | {Response #1}=<br>1. Fracture of the silk fibers due to tensile strain<br>2. Delamination between the silk and metal layers<br>3. Debonding between the silk and carbon nanotubes<br>4. Fracture of the carbon nanotubes<br>5. Deformation of the metal components<br><br>These failure mechanisms consider the different materials present in the hybrid nanocomposite and take into account their individual properties and interactions. | |
| Read this: '{Response #1}'. List the most important concepts to answer the question '{question}'. | {Response #2}=<br>The most important concepts to answer the question 'What would be likely failure mechanisms of a hybrid silk-metal nanocomposite, reinforced with carbon nanotubes?' are fracture of the silk fibers due to tensile strain, delamination between the silk and metal layers, debonding between the silk and carbon nanotubes, fracture of the carbon nanotubes, and deformation of the metal components. | Here we use the LLM to identify the most critical concepts to answer the question. |
| Considering '{Response #2}', answer this question with a detailed response: {question} | The likely failure mechanisms of a hybrid silk-metal nanocomposite, reinforced with carbon nanotubes, include fracture of the silk fibers due to tensile strain, delamination between the silk and metal layers, debonding between the silk and carbon nanotubes, fracture of the carbon nanotubes, and deformation of the metal components. | |
| Single-shot answer to the question. | The likely failure mechanisms of a hybrid silk-metal nanocomposite, reinforced with carbon nanotubes, would be determined by the interplay between the silk and metal components. The silk fibers, being natural materials, are susceptible to biodegradation and environmental factors. The metal nanoparticles, on the other hand, provide strength and durability but may also be prone to corrosion and wear. The carbon nanotubes, being carbon-based, can add stability and conductivity to the material but may also be affected by environmental factors. Overall, the failure mechanisms would depend on the specific conditions and interactions between the silk, metal, and carbon nanotubes. | Answer is more general, lacks details on mechanisms. |

**Table 7**: Multi-step sampling inspired by the Tree-of-Thought method, implemented via variegated uses of the LLM as generator of ideas/concepts and to judge these against their usefulness to solves a particular task. The question asked is: "**The mechanism of flaw-tolerance in alpha-helical protein networks is: "The geometric transformation of the crack shape is believed to help minimize stress concentration at corners, although the exact mechanism is not fully understood. As the crack undergoes a series of changes, from mode I to a circular hole, and eventually becoming an elongated crack aligned with the direction of loading, the stress concentration at corners is reduced. This transformation occurs as individual proteins within the crack continuously unfold, independent of their neighboring proteins. The reduced stresses in the vicinity of the elongated crack contribute to a decrease in stress concentration at corners. Overall, this geometric transformation of the crack shape is thought to provide self-protection and distribute stress more evenly along the crack, ultimately minimizing stress concentration." Now, describe how to translate this concept to a material composed out of thin glass fibers as elementary building blocks. Define details of how the glass fibers must be engineered, and how they are arranged into larger-scale structures."** We proceed by asking several questions to sample dynamically and iteratively by using MechGPT to judge and assess its own responses. Compared to the single-shot answer, the complex prompting strategy elicits more nuanced and detailed responses. It is noted that in this example we do not use RAG, and only the question designated above is used to prompt the model.

| **Question** | **Response** | **Notes** |
|---|---|---|
| Generate a list of initial thoughts that are relevant for answering this question: '{question}'. Do not answer the question. | {Response #1}=<br>Sure, here are some initial thoughts relevant to answering the question:<br><br>- The mechanism of flaw-tolerance in alpha-helical protein networks is believed to involve the geometric transformation of the crack shape, which helps minimize stress concentration at corners.<br>- As the crack undergoes changes, the stress concentration at corners is reduced, and the stress is distributed more evenly along the crack.<br>- Individual proteins within the crack unfold independently, contributing to the reduction of stress concentration.<br>- The overall effect is a self-protection mechanism that distributes stress more evenly, minimizing stress concentration.<br><br>To apply this concept to a material composed out of thin glass fibers as elementary building blocks, the glass fibers must be engineered to have specific properties. The arrangement of the glass fibers into larger-scale structures is crucial to ensure the desired strength and flaw-tolerance. The exact engineering details may need to be determined based on the specific requirements of the material. | In the first step we ask MechGPT to generate a set of initial concepts relevant to the query. |
| Read this: '{Response #1}'. List the most important concepts to answer the question '{question}'. | {Response #2}=<br>The mechanism of flaw-tolerance in alpha-helical protein networks is believed to involve the geometric transformation of the crack shape, which helps minimize stress concentration at corners. As the crack undergoes changes, the stress concentration at corners is reduced, and the stress is distributed more evenly along the crack. Individual proteins within the crack unfold independently, contributing to the reduction of stress concentration. The overall effect is a self-protection mechanism that distributes stress more evenly, minimizing stress concentration.<br><br>To apply this concept to a material composed out of thin glass fibers as elementary building blocks, the glass fibers must be engineered to have specific properties. The arrangement of the glass fibers into larger-scale structures is crucial to ensure the desired strength and flaw-tolerance. The exact engineering details may need to be determined based on the specific requirements of the material. | Here we use the LLM to identify the most critical concepts to answer the question. |
| Considering '{Response #2}', answer this question with a | To apply this concept to a material composed out of thin glass fibers as elementary building blocks, the glass fibers must be engineered to have specific properties. The arrangement of the glass fibers into larger-scale structures is crucial to ensure the desired strength and flaw-tolerance. The exact engineering | Now we prompt the LLM to answer the question. Key issues reproduced agree with our |
| detailed response: {question} | details may need to be determined based on the specific requirements of the material. | understanding of the key aspects, including larger-scale mesh structures, and the significance of ensuring that the fibers themselves have specific properties (akin to the protein filaments that can stretch exhaustively without breaking, as laid out in [65,108]). |
| Single-shot answer to the question. | Sure! What is the question regarding the engineering of glass fibers as elementary building blocks in a material? | No answer provided. Alternative responses may include hallucinated results. |

**Table 8**: Agent-based modeling, applied here to solve a molecular design problem. The set of agents include: The "User" (human that poses the question), a "Planner" that develops a plan to answer the question, a "Coordinate retriever" that can retrieve coordinates from SMILES coordinates or its own knowledge, and a "Chatbot" with broad skills and execution power. The agents have access to a Python function that converts SMILES code into atom types and molecular coordinates (the molecular structure), as well as a Python function that conducts a DFT simulation given a molecular structure as input. The question posed here is: "**Consider this molecule in SMILES representation: CCCCXCC, where X is one of (C, O, N). Which of these options leads to the lowest energy structure?**" The agents devise a plan, carry out the simulations, analyze the data and provide the final response to the User.

| **Conversation** | **Notes** |
|---|---|
| **User**: Consider this molecule in SMILES representation: CCCCXCC, where X is one of (C, O, N). Which of these options leads to the lowest energy structure? | Initial question posed |
| **Planner**: Sure, to solve this task we need to convert the SMILES representation into coordinates for each molecule, calculate the energy for each molecule, and then compare the energy level.<br><br>Let's detail this in a precise plan:<br><br>1. For each of the atoms C, O, N as X:<br>    - Replace X in the SMILES "CCCCXCC" with the atom.<br>    - Use the function \`functions.coords_from_SMILES\` with the new SMILES representation to get the coordinates of the molecule.<br>    - Use the function \`functions.query_DFT\` with these coordinates to calculate the energy of the molecule.<br>2. Compare the calculated energy levels for the three molecules.<br>3. The atom that results in the lowest energy level molecule is the one that should replace X. | Planner develops a strategy to solve the problem |
| **Coordinate retriever**: Understood. Let's get started with step 1, where we'll generate coordinates for each of the molecules and calculate their energies.<br>***** Suggested function Call: coords_from_SMILES *****<br>Arguments:<br>{<br>  "SMILES": "CCCCCCC"<br>} | The other agents follow the plan and retrieve information (starting with coordinate retrieval for each of the structures, energy calculation, etc.); repeated for each of the three cases of X=(C, O, N). |
| >>>>>>>> EXECUTING FUNCTION coords_from_SMILES...<br>**Chatbot**:<br>***** Response from calling function "coords_from_SMILES" *****<br>['C 2.459 0.3911 0.8023', 'C 2.528 -0.475 -0.4535', 'C 1.304 -<br>0.309 -1.369', 'C -0.006061 -0.8418 -0.7517', 'C -0.8906 0.2868 -<br>0.2008', 'C -2.163 -0.2759 0.4426', 'C -3.05 0.8428 0.9815', 'H<br>1.67 0.01777 1.488', 'H 2.254 1.45 0.5352', 'H 3.43 0.3431<br>1.338', 'H 2.644 -1.542 -0.1625', 'H 3.436 -0.1851 -1.027', 'H<br>1.203 0.7551 -1.677', 'H 1.508 -0.8856 -2.298', 'H 0.2104 -1.583<br>0.04887', 'H -0.5783 -1.375 -1.543', 'H -1.172 0.9684 -1.034', 'H -<br>0.3301 0.8759 0.5549', 'H -1.891 -0.9536 1.281', 'H -2.738 -0.8594<br>-0.3097', 'H -3.962 0.4059 1.441', 'H -2.507 1.426 1.755', 'H -<br>3.358 1.522 0.158']<br>**********************************************************************<br><br>***** Suggested function Call: query_DFT *****<br>Arguments:<br>{<br>  "coordinates": "C 2.459 0.3911 0.8023\nC 2.528 -0.475 -0.4535\nC<br>1.304 -0.309 -1.369\nC -0.006061 -0.8418 -0.7517\nC -0.8906<br>0.2868 -0.2008\nC -2.163 -0.2759 0.4426\nC -3.05 0.8428<br>0.9815\nH 1.67 0.01777 1.488\nH 2.254 1.45 0.5352\nH 3.43<br>0.3431 1.338\nH 2.644 -1.542 -0.1625\nH 3.436 -0.1851 -<br>1.027\nH 1.203 0.7551 -1.677\nH 1.508 -0.8856 -2.298\nH<br>0.2104 -1.583 0.04887\nH -0.5783 -1.375 -1.543\nH -1.172 0.9684<br>-1.034\nH -0.3301 0.8759 0.5549\nH -1.891 -0.9536 1.281\nH -<br>2.738 -0.8594 -0.3097\nH -3.962 0.4059 1.441\nH -2.507 1.426<br>1.755\nH -3.358 1.522 0.158"<br>}<br>>>>>>>>> EXECUTING FUNCTION query_DFT... | Get coordinates of CCCCCCC, and run DFT |
| […..]<br>E1 = -997.6630956252707 E_coul = 406.0492308406011<br>Extra cycle E= -274.409942204028 delta_E= -5.68e-12 \|g\|= 1.45e-06 \|ddm\|= 5.38e-06<br>CPU time for scf_cycle 59.21 sec, wall time 3.03 sec<br>CPU time for SCF 59.28 sec, wall time 3.05 sec<br>converged SCF energy = -274.409942204028<br><br>***** Response from calling function "query_DFT" *****<br>-274.4099422040276 | |
| **Coordinate retriever**:<br>***** Suggested function Call: coords_from_SMILES *****<br>Arguments:<br>{<br>"SMILES": "CCCCOCC"<br>} | Get coordinates of CCCCOCC, and run DFT |
| >>>>>>>> EXECUTING FUNCTION coords_from_SMILES...<br>**Chatbot**:<br>***** Response from calling function "coords_from_SMILES" *****<br>['C 3.481 0.4249 -0.2857', 'C 2.112 0.8256 0.255', 'C 1.052 -0.2277 -0.08758', 'C -0.3203 0.172 0.4532', 'O -1.257 -0.8308 0.1181', 'C -2.564 -0.4335 0.4876', 'C -3.297 0.1551 -0.7131', 'H 3.445 0.3256 -1.392', 'H 3.805 -0.5403 0.1593', 'H 4.227 1.204 -0.02324', 'H 1.824 1.805 -0.1858', 'H 2.181 0.9428 1.359', 'H 0.9859 -0.3436 -1.192', 'H 1.344 -1.206 0.3525', 'H -0.2527 0.275 1.56', 'H -0.6083 1.151 0.005801', 'H -2.594 0.285 1.341', 'H -3.116 -1.338 0.8194', 'H -3.305 -0.5765 -1.549', 'H -2.798 1.087 -1.051', 'H -4.344 0.3929 -0.4327']<br>****************************************************************<br><br>***** Suggested function Call: query_DFT *****<br>Arguments:<br>{<br>"coordinates": "C 3.481 0.4249 -0.2857\nC 2.112 0.8256 0.255\nC 1.052 -0.2277 -0.08758\nC -0.3203 0.172 0.4532\nO -1.257 -0.8308 0.1181\nC -2.564 -0.4335 0.4876\nC -3.297 0.1551 -0.7131\nH 3.445 0.3256 -1.392\nH 3.805 -0.5403 0.1593\nH 4.227 1.204 -0.02324\nH 1.824 1.805 -0.1858\nH 2.181 0.9428 1.359\nH 0.9859 -0.3436 -1.192\nH 1.344 -1.206 0.3525\nH -0.2527 0.275 1.56\nH -0.6083 1.151 0.005801\nH -2.594 0.285 1.341\nH -3.116 -1.338 0.8194\nH -3.305 -0.5765 -1.549\nH -2.798 1.087 -1.051\nH -4.344 0.3929 -0.4327"<br>}<br>**********************************************<br><br>>>>>>>>> EXECUTING FUNCTION query_DFT...<br>[…]<br>E1 = -1047.5752594421979 E_coul = 420.8765995972293<br>Extra cycle E= -310.227522287807 delta_E= -3.98e-12 \|g\|= 1.16e-06 \|ddm\|= 5.03e-06<br>CPU time for scf_cycle 68.75 sec, wall time 3.51 sec<br>CPU time for SCF 68.86 sec, wall time 3.54 sec<br>converged SCF energy = -310.227522287807<br><br>***** Response from calling function "query_DFT" *****<br>-310.22752228780735 | |
| **Coordinate retriever**:<br>***** Suggested function Call: coords_from_SMILES *****<br>Arguments:<br>{<br>"SMILES": "CCCCNCC"<br>}<br>>>>>>>>> EXECUTING FUNCTION coords_from_SMILES...<br>***** Response from calling function "coords_from_SMILES" *****<br>['C -3.667 0.4902 -0.2323', 'C -2.421 -0.338 -0.5355', 'C -1.151 0.3803 -0.06634', 'C 0.09811 -0.4524 -0.3695', 'N 1.299 0.269 0.06676', 'C 2.505 -0.4832 -0.298', 'C 3.751 0.3111 0.07632', 'H -3.758 0.6629 0.8612', 'H -4.569 -0.05361 -0.5827', 'H -3.616 1.469 | Get coordinates of CCCCNCC, and run DFT |
| -0.7556', 'H -2.509 -1.32 -0.02114', 'H -2.368 -0.518 -1.631', 'H -1.213 0.5621 1.029', 'H -1.068 1.362 -0.5827', 'H 0.1453 -0.6358 -1.467', 'H 0.02723 -1.432 0.1556', 'H 1.269 0.3373 1.112', 'H 2.527 -1.464 0.2291', 'H 2.526 -0.665 -1.396', 'H 3.784 0.4793 1.174', 'H 3.748 1.292 -0.4442', 'H 4.659 -0.2529 -0.2257'] | |
| **Chatbot**:<br>***** Suggested function Call: query_DFT *****<br>Arguments:<br>{<br>"coordinates": "C -3.667 0.4902 -0.2323\nC -2.421 -0.338 -0.5355\nC -1.151 0.3803 -0.06634\nC 0.09811 -0.4524 -0.3695\nN 1.299 0.269 0.06676\nC 2.505 -0.4832 -0.298\nC 3.751 0.3111 0.07632\nH -3.758 0.6629 0.8612\nH -4.569 -0.05361 -0.5827\nH -3.616 1.469 -0.7556\nH -2.509 -1.32 -0.02114\nH -2.368 -0.518 -1.631\nH -1.213 0.5621 1.029\nH -1.068 1.362 -0.5827\nH 0.1453 -0.6358 -1.467\nH 0.02723 -1.432 0.1556\nH 1.269 0.3373 1.112\nH 2.527 -1.464 0.2291\nH 2.526 -0.665 -1.396\nH 3.784 0.4793 1.174\nH 3.748 1.292 -0.4442\nH 4.659 -0.2529 -0.2257"<br>}<br>**********************************************<br><br>>>>>>>>> EXECUTING FUNCTION query_DFT...<br>[..]<br>E1 = -1008.918528295655 E_coul = 407.03364174745474<br>Extra cycle E= -290.399363184172 delta_E= -1.63e-11 \|g\|= 2.25e-06 \|ddm\|= 1.09e-05<br>CPU time for scf_cycle 63.95 sec, wall time 3.29 sec<br>CPU time for SCF 64.02 sec, wall time 3.31 sec<br>converged SCF energy = -290.399363184172<br><br>***** Response from calling function "query_DFT" *****<br>-290.39936318417176 | |
| **Planner**: The energies of the molecules with C, O and N are -274.41, -310.23 and -290.4 respectively. Thus, the molecule with O (CCCCOCC) leads to the lowest energy structure. | The Planner considers all simulation results, analyzes and answers the question. |
| **Chatbot**: The molecule CCCCCOCC has the lowest energy structure. TERMINATE | The Chatbot provides the final answer after checking the results. |

**Table 9**: Agent-based modeling, applied here to answer a complex design question: "*Design molybdenene 2D materials by using ideas of flaw-tolerance in alpha-helical protein meshes. Develop three design principles that incorporate important mechanisms.*". The set of agents include: The "Boss" (human that poses the question), a "Senior Engineer" that develops a plan to answer the question, a "Modeling Expert", and a "Reviewer" who critically assesses the responses and integrates knowledge. The boss is the human user, who asks the questions and gives approval and input at different stages of the conversation. For instance, this includes approval of the plan, and follow-up questions. Human input in the conversation below is marked in *italic*. Each of the agents can use experts with deep technical knowledge for additional information (see **Figure 7** for a flowchart and overview of the approach). Below is a summary of how the conversation unfolds.

| **Conversation** | **Notes** |
|---|---|
| **Boss**: Design molybdenene 2D materials by using ideas of flaw-tolerance in alpha-helical protein meshes. Develop three design principles that incorporate important mechanisms. | Initial question is posed by the Boss, on behalf of the human user. |
| **Senior Engineer**: Sure, here is the plan:<br><br>1. **Researcher's task**: The researcher will use the `retrieve_content_protein` function to gather information about the flaw-tolerance in alpha-helical protein meshes. They will also use the `retrieve_content_moly` function to gather information about molybdenene 2D materials. The researcher will then synthesize this information to develop three design principles that incorporate important mechanisms.<br><br>2. **Modeling Expert's task**: The modeling expert will use the `retrieve_content_book` function to gather information about atomistic and multiscale modeling, especially materials deformation and failure. They will then use this information to suggest suitable modeling methods for the design of molybdenene 2D materials based on the principles developed by the researcher.<br><br>3. **Reviewer's task**: The reviewer will critically evaluate the design principles developed by the researcher and the modeling methods suggested by the modeling expert. They will provide feedback and suggest improvements if necessary.<br><br>After these steps, I will consolidate all the information and present it to you for approval.<br><br>Does this plan meet your expectations, Boss? | The Senior Engineer develops a plan including detailed retrieval strategies for key information.<br><br>As instructed, the Boss is asked to review and approve the plan. |
| **Boss**: *This plan is excellent. Please proceed to answer the question.* | The human user approves the plan (potentially, the human user can provide alternative suggestions or critically assess, in which case the Senior Engineer would revise the plan). |
| **Senior Engineer**:<br>***** Suggested function Call: retrieve_content_protein *****<br>Arguments:<br>{<br>"message": "flaw-tolerance in alpha-helical protein meshes"<br>}<br>*********************************************************<br>>>>>>>>> EXECUTING FUNCTION retrieve_content_protein...<br>doc_ids: [['doc_13', 'doc_2']]<br>Adding doc_id doc_13 to context.<br>Adding doc_id doc_2 to context.<br>Modeling_expert (to chat_manager):<br>***** Response from calling function "retrieve_content_protein" *****<br>[….retrieves chunks of text from the protein knowledge database…]<br><br>***** Suggested function Call: retrieve_content_moly *****<br>Arguments:<br>{<br>"message": "molybdenene 2D materials"<br>} | The Senior Engineer begins to query the expert agents – first, the protein expert assistant, then the molybdenene expert assistant. |
| >>>>>>>> EXECUTING FUNCTION retrieve_content_moly...<br>doc_ids:  [['doc_1', 'doc_8']]<br>Adding doc_id doc_1 to context.<br>Adding doc_id doc_8 to context.<br>Modeling_expert (to chat_manager):<br>***** Response from calling function "retrieve_content_moly" *****<br>[….retrieves chunks of text from the molybdene dataset…] | |
| **Modeling expert:**<br>***** Suggested function Call: retrieve_content_book *****<br>Arguments:<br>{<br>  "message": "atomistic and multiscale modeling, materials deformation and failure"<br>}<br>>>>>>>>> EXECUTING FUNCTION retrieve_content_book...<br>doc_ids:  [['doc_9', 'doc_492']]<br>Adding doc_id doc_9 to context.<br>Adding doc_id doc_492 to context.<br>Reviewer (to chat_manager):<br>[….retrieves chunks of text from the book dataset…] | The modeling expert receives feedback from the modeling expert assistant. |
| **Reviewer**: Based on the information gathered, here are three design principles for molybdenene 2D materials using ideas of flaw-tolerance in alpha-helical protein meshes:<br><br>1. **Energy Dissipation**: Alpha-helical protein meshes have a high energy dissipation ability due to the structure of alpha-helical proteins which combine a large array of small groups of H-bonds. These unfold concurrently in groups of 3-4 at relatively small force levels, providing a strongly nonlinear material behavior. This principle can be applied to molybdenene 2D materials by designing them in a way that allows for high energy dissipation, perhaps through the incorporation of similar small groups of bonds that can unfold under stress.<br><br>2. **Self-Healing**: The use of H-bonds in alpha-helical protein meshes renders the structure self-healing, as H-bonds can reform at moderate temperatures and restore the initial structure even after severe deformation. This principle can be applied to molybdenene 2D materials by incorporating elements that can reform after deformation, thereby increasing the material's durability and lifespan.<br><br>3. **Flaw-Tolerance**: Alpha-helical protein meshes do not display a strong stress concentration at corners of cracks, which is a common cause of failure in conventional solids. This is due to the ability of each filament in the mesh to independently stretch without affecting neighboring bonds. This principle can be applied to molybdenene 2D materials by designing them in a way that minimizes stress concentration at potential points of failure.<br><br>To model these principles, multiscale modeling techniques can be used. These techniques allow a rigorous linking of material properties from quantum mechanics to mesoscale and macroscale, which could be crucial in designing molybdenene 2D materials that incorporate the principles derived from alpha-helical protein meshes.<br><br>In addition, the use of computational experiments can help in the design and testing of these materials. For instance, the behavior of the materials under different stress conditions can be simulated to ensure that they adhere to the principles of energy dissipation, self-healing, and flaw-tolerance.<br><br>Finally, experimental techniques such as TEM and nanoindentation can be used to analyze the structure and mechanical properties of the designed materials. These techniques can provide valuable insights into the effectiveness of the design principles and can guide further improvements. | |
| **Boss**: *Thank you. Can you elaborate a bit more on how flaw-tolerance can be achieved in molybdenene, especially in terms of structural design?* | Human user input is sought, with the possibility to add additional questions. Here, the human user asks for additional details on one of the aspects of the answer. |
| **Senior Engineer**: Certainly, Boss. | The Senior Engineer provides additional details, as requested by the human user. |
| Flaw-tolerance in molybdenene 2D materials can be achieved by borrowing the concept of independent stretching of filaments from alpha-helical protein meshes. In these protein meshes, each filament can stretch independently without affecting neighboring bonds. This is possible due to the absence of immediate interactions between individual filaments in the network, aside from cross-links at node points of the mesh. This design allows the network to avoid strong stress concentration at the corners of cracks, which is a common cause of failure in conventional solids.<br><br>In the context of molybdenene 2D materials, this principle can be applied by designing the material in a way that allows for independent movement of its constituent parts under stress. This could potentially be achieved by incorporating a network-like structure into the material, similar to the mesh structure of alpha-helical protein networks.<br><br>In this network, the constituent parts of the molybdenene material could be designed to move independently under stress, thereby dissipating energy and avoiding the concentration of stress at any particular point. This would require careful consideration of the bonds between the constituent parts of the material, to ensure that they allow for independent movement while still maintaining the overall integrity of the material.<br><br>Furthermore, the use of atomistic and multiscale modeling can help in designing such a structure. These modeling techniques can provide detailed insights into the behavior of the material at different scales, from the atomic level to the macroscopic level. This can guide the design process and help in achieving a flaw-tolerant structure.<br><br>I hope this provides a clearer picture of how flaw-tolerance can be achieved in molybdenene 2D materials. If you have any further questions, please let me know. | |
| **Boss**: *Thank you. Can you comment on how such a network-like structure could be manufactured in molybdenene?* | Another follow-up question is posed by the human user. |
| **Modeling expert**: Manufacturing a network-like structure in molybdenene 2D materials could be a challenging task due to the atomic scale of the material. However, advanced nanotechnology techniques could potentially be used to achieve this.<br><br>One possible method could be the use of a bottom-up approach, such as chemical vapor deposition (CVD), where individual atoms or molecules are assembled to form the desired structure. In the case of molybdenene, this could involve the controlled deposition of molybdenum atoms in a way that forms a network-like structure. This would require precise control over the deposition process to ensure that the atoms are arranged in the desired pattern.<br><br>Another possible method could be the use of a top-down approach, such as lithography or etching, where a bulk material is carved to create the desired structure. This could involve the use of advanced lithography techniques to carve a network-like structure into a bulk molybdenene material. However, this method could be more challenging due to the difficulty of carving such a fine structure into the material.<br><br>Regardless of the method used, the manufacturing process would likely require the use of advanced characterization techniques, such as transmission electron microscopy (TEM), to verify the structure of the material. Additionally, computational modeling could play a crucial role in guiding the manufacturing process and predicting the behavior of the material.<br><br>It's important to note that these are just potential methods and the actual feasibility would depend on various factors, including the specific properties of molybdenene and the available manufacturing technology. | The Modeling Expert provides an answer. |
| **Boss**: Thank you! | The conversation ends once the human user approves. |

**Table 10**: Agent-based modeling, applied here to write code, execute and plot the results. The task given is: "Use PySCF to calculate the energy of a O2 molecule, for varying bond lengths between 0.7 and 1.8 Angstroms in increments of 0.1 Angstroms. Fit a piecewise spline curve to the data. Then plot the energy over the bond length for the data and the fitted potential, and save as plot_O2_spline_fit_potential.svg. Save the parameters of the spline curve fit into a file, in JSON format. Save the calculation results in a file, in JSON format. What is the lowest energy configuration?" As can be verified in the conversation, the set of agents not only write code, install any required libraries, and execute it, but also correct any errors that may occur (this self-correcting behavior is key to achieve successful complex code development and exection).

<table>
<tr><th>Conversation</th><th>Notes</th></tr>
<tr><td>User: Use PySCF to calculate the energy of a O2 molecule, for varying bond lengths betwe en 0.7 and 1.8 Angstroms in increments of 0.1 Angstroms. Fit a piecewise spline curve to t he data. Then plot the energy over the bond length for the data and the fitted potential, and save as plot_O2_spline_fit_potential.svg. Save the parameters of the spline curve fit into a f ile, in JSON format. Save the calculation results in a file, in JSON format. What is the lowe st energy configuration?</td><td>The User posts the tasks and initi ates the conversation.</td></tr>
<tr><td>Assistant: To calculate the energy of an O2 molecule using PySCF and fit a piecewise spli ne curve to the data, we can follow these steps:<br><br>1. Install the necessary packages:<br>  - PySCF: `pip install pyscf`<br>  - SciPy: `pip install scipy`<br>  - NumPy: `pip install numpy`<br>  - Matplotlib: `pip install matplotlib`<br><br>2. Import the required modules:<br><br>```python<br>import numpy as np<br>from scipy.interpolate import CubicSpline<br>import matplotlib.pyplot as plt<br>from pyscf import gto, dft<br>import json<br>```<br><br>3. Define the bond lengths and an empty list to store the energy values:<br><br>```python<br>bond_lengths = np.arange(0.7, 1.9, 0.1)<br>energies = []<br>```<br><br>4. Perform the calculations for each bond length:<br><br>```python<br>for bond_length in bond_lengths:<br>    mol = gto.M(atom='O 0 0 0; O 0 0 {}'.format(bond_length), basis='sto-3g')<br>    mf = dft.RKS(mol)<br>    mf.xc = 'pbe'<br>    energy = mf.kernel()<br>    energies.append(energy)<br>```<br><br>5. Fit a piecewise spline curve to the data:<br><br>```python<br>spline = CubicSpline(bond_lengths, energies)<br>```<br><br>6. Plot the energy over the bond length for the data and the fitted potential:<br><br>```python<br>plt.plot(bond_lengths, energies, 'o', label='Data')</td><td>The Assistant provides the first r esponse, including a plan to insta ll packages, import models and i ntegrating it as a code for execut ion.</td></tr>
</table>

<table>
<tr>
<td>plt.plot(bond_lengths, spline(bond_lengths), label='Spline Fit')<br>plt.xlabel('Bond Length (Å)')<br>plt.ylabel('Energy (Hartree)')<br>plt.legend()<br>plt.savefig('plot_O2_spline_fit_potential.svg')<br>plt.show()<br>```<br><br>7. Save the parameters of the spline curve fit into a file in JSON format:<br><br>```python<br>spline_params = {'knots': spline.x.tolist(), 'coefficients': spline.c.tolist(), 'degree': spline.k}<br>with open('spline_params.json', 'w') as f:<br>    json.dump(spline_params, f)<br>```<br><br>8. Save the calculation results in a file in JSON format:<br><br>```python<br>calculation_results = {'bond_lengths': bond_lengths.tolist(), 'energies': energies}<br>with open('calculation_results.json', 'w') as f:<br>    json.dump(calculation_results, f)<br>```<br><br>To find the lowest energy configuration, we can use the `min` function to find the minimu m energy value and its corresponding bond length:<br><br>```python<br>min_energy = min(energies)<br>min_bond_length = bond_lengths[np.argmin(energies)]<br>```<br><br>Finally, print the lowest energy configuration:<br><br>```python<br>print('The lowest energy configuration is at a bond length of {} Å with an energy of {} Har tree.'.format(min_bond_length, min_energy))<br>```<br><br>Please note that the code assumes you have the necessary packages installed and that you h ave the required permissions to save files in the current directory.</td>
<td></td>
</tr>
<tr>
<td>>>>>>>>> EXECUTING CODE BLOCK 0 (inferred language is python)...<br><br>>>>>>>>> EXECUTING CODE BLOCK 1 (inferred language is python)...<br>User:<br>exitcode: 1 (execution failed)<br>Code output:<br><br>Traceback (most recent call last):<br>  File "", line 1, in <module><br>    bond_lengths = np.arange(0.7, 1.9, 0.1)<br>NameError: name 'np' is not defined</td>
<td>The User agent tries to run the c ode; however, error messages ar e retrieved.</td>
</tr>
<tr>
<td>Assistant: It seems that the `numpy` module (`np`) is not imported. Let's modify the code t o include the necessary import statement for `numpy`:<br><br>```python<br>import numpy as np<br>from scipy.interpolate import CubicSpline<br>import matplotlib.pyplot as plt<br>from pyscf import gto, dft<br>import json<br><br>bond_lengths = np.arange(0.7, 1.9, 0.1)<br>energies = []</td>
<td>The agent considers the error me ssages and revises the code base d on the feedback.</td>
</tr>
</table>

| | |
|---|---|
| for bond_length in bond_lengths:<br>    mol = gto.M(atom='O 0 0 0; O 0 0 {}'.format(bond_length), basis='sto-3g')<br>    mf = dft.RKS(mol)<br>    mf.xc = 'pbe'<br>    energy = mf.kernel()<br>    energies.append(energy)<br><br>spline = CubicSpline(bond_lengths, energies)<br><br>plt.plot(bond_lengths, energies, 'o', label='Data')<br>plt.plot(bond_lengths, spline(bond_lengths), label='Spline Fit')<br>plt.xlabel('Bond Length (Å)')<br>plt.ylabel('Energy (Hartree)')<br>plt.legend()<br>plt.savefig('plot_O2_spline_fit_potential.svg')<br>plt.show()<br><br>spline_params = {'knots': spline.x.tolist(), 'coefficients': spline.c.tolist(), 'degree': spline.k}<br>with open('spline_params.json', 'w') as f:<br>    json.dump(spline_params, f)<br><br>calculation_results = {'bond_lengths': bond_lengths.tolist(), 'energies': energies}<br>with open('calculation_results.json', 'w') as f:<br>    json.dump(calculation_results, f)<br><br>min_energy = min(energies)<br>min_bond_length = bond_lengths[np.argmin(energies)]<br><br>print('The lowest energy configuration is at a bond length of {} Å with an energy of {} Har tree.'.format(min_bond_length, min_energy))<br>```<br><br>Please try running the modified code again. Let me know if you encounter any further issue s. | |
| >>>>>>>> EXECUTING CODE BLOCK 0 (inferred language is python)...<br>**User** :<br>exitcode: 1 (execution failed)<br>Code output:<br>/home/mbuehler/miniconda3/envs/PyTorch/lib/python3.8/site-packages/pyscf/dft/libxc.py: 771: UserWarning: Since PySCF-2.3, B3LYP (and B3P86) are changed to the VWN-RPA variant, corresponding to the original definition by Stephens et al. (issue 1480) and the sam e as the B3LYP functional in Gaussian. To restore the VWN5 definition, you can put the se tting "B3LYP_WITH_VWN5 = True" in pyscf_conf.py<br>  warnings.warn('Since PySCF-2.3, B3LYP (and B3P86) are changed to the VWN-RPA var iant, '<br>Traceback (most recent call last):<br>  File "", line 27, in <module><br>    spline_params = {'knots': spline.x.tolist(), 'coefficients': spline.c.tolist(), 'degree': spline.k }<br>AttributeError: 'CubicSpline' object has no attribute 'k' | The User executes the revised, u pdated code. Another error mess age is received. |
| **Assistant**: I apologize for the confusion. The `CubicSpline` object in SciPy does not have a `k` attribute. The degree of the spline is always 3 for `CubicSpline`, so we don't need to sav e it. Here is the corrected code:<br><br>```python<br>import numpy as np<br>from scipy.interpolate import CubicSpline<br>import matplotlib.pyplot as plt<br>from pyscf import gto, dft<br>import json<br><br>bond_lengths = np.arange(0.7, 1.9, 0.1)<br>energies = [] | The Assistant again consider the error messages and further revise s the code. Explanations are prov ided as to possible strategies to a ddress the issue. |
| for bond_length in bond_lengths:<br>    mol = gto.M(atom='O 0 0 0; O 0 0 {}'.format(bond_length), basis='sto-3g')<br>    mf = dft.RKS(mol)<br>    mf.xc = 'pbe'<br>    energy = mf.kernel()<br>    energies.append(energy)<br><br>spline = CubicSpline(bond_lengths, energies)<br><br>plt.plot(bond_lengths, energies, 'o', label='Data')<br>plt.plot(bond_lengths, spline(bond_lengths), label='Spline Fit')<br>plt.xlabel('Bond Length (Å)')<br>plt.ylabel('Energy (Hartree)')<br>plt.legend()<br>plt.savefig('plot_O2_spline_fit_potential.svg')<br>plt.show()<br><br>spline_params = {'knots': spline.x.tolist(), 'coefficients': spline.c.tolist()}<br>with open('spline_params.json', 'w') as f:<br>    json.dump(spline_params, f)<br><br>calculation_results = {'bond_lengths': bond_lengths.tolist(), 'energies': energies}<br>with open('calculation_results.json', 'w') as f:<br>    json.dump(calculation_results, f)<br><br>min_energy = min(energies)<br>min_bond_length = bond_lengths[np.argmin(energies)]<br><br>print('The lowest energy configuration is at a bond length of {} Å with an energy of {} Hartree.'.format(min_bond_length, min_energy))<br>```<br><br>Please try running the modified code again. Let me know if you encounter any further issues. | |
| >>>>>>>> EXECUTING CODE BLOCK 0 (inferred language is python)...<br>**User**:<br>exitcode: 0 (execution succeeded)<br>Code output:<br>converged SCF energy = -145.62788167137<br>converged SCF energy = -146.846811710288<br>converged SCF energy = -147.494980044753<br>converged SCF energy = -147.830098170465<br>converged SCF energy = -147.994263551652<br>converged SCF energy = -148.065052309465<br>converged SCF energy = -148.084209872691<br>converged SCF energy = -148.074335875387<br>converged SCF energy = -148.048364221058<br>converged SCF energy = -148.014377520708<br>converged SCF energy = -147.977673841903<br>converged SCF energy = -147.941590421713<br>Figure(640x480)<br>The lowest energy configuration is at a bond length of 1.2999999999999998 Å with an energy of -148.08420987269093 Hartree. | The User executes the updated code, with successful completion. |
| **Assistant:** Great! The code executed successfully.<br><br>The lowest energy configuration for the O2 molecule is at a bond length of approximately 1.3 Å with an energy of -148.08420987269093 Hartree.<br><br>The plot of the energy over the bond length for the data and the fitted potential has been saved as 'plot_O2_spline_fit_potential.svg'. | The Assistant uses the result of the code execution to answer additional questions (here, specifically the bond length at minimum energy, the stable configuration). It also provide a summary of all files/data produced, including the JSON files. |
| The parameters of the spline curve fit have been saved in a file named 'spline_params.json' in JSON format.<br><br>The calculation results have been saved in a file named 'calculation_results.json' in JSON f ormat.<br><br>Please check these files in your current directory. | |

TOC graphic

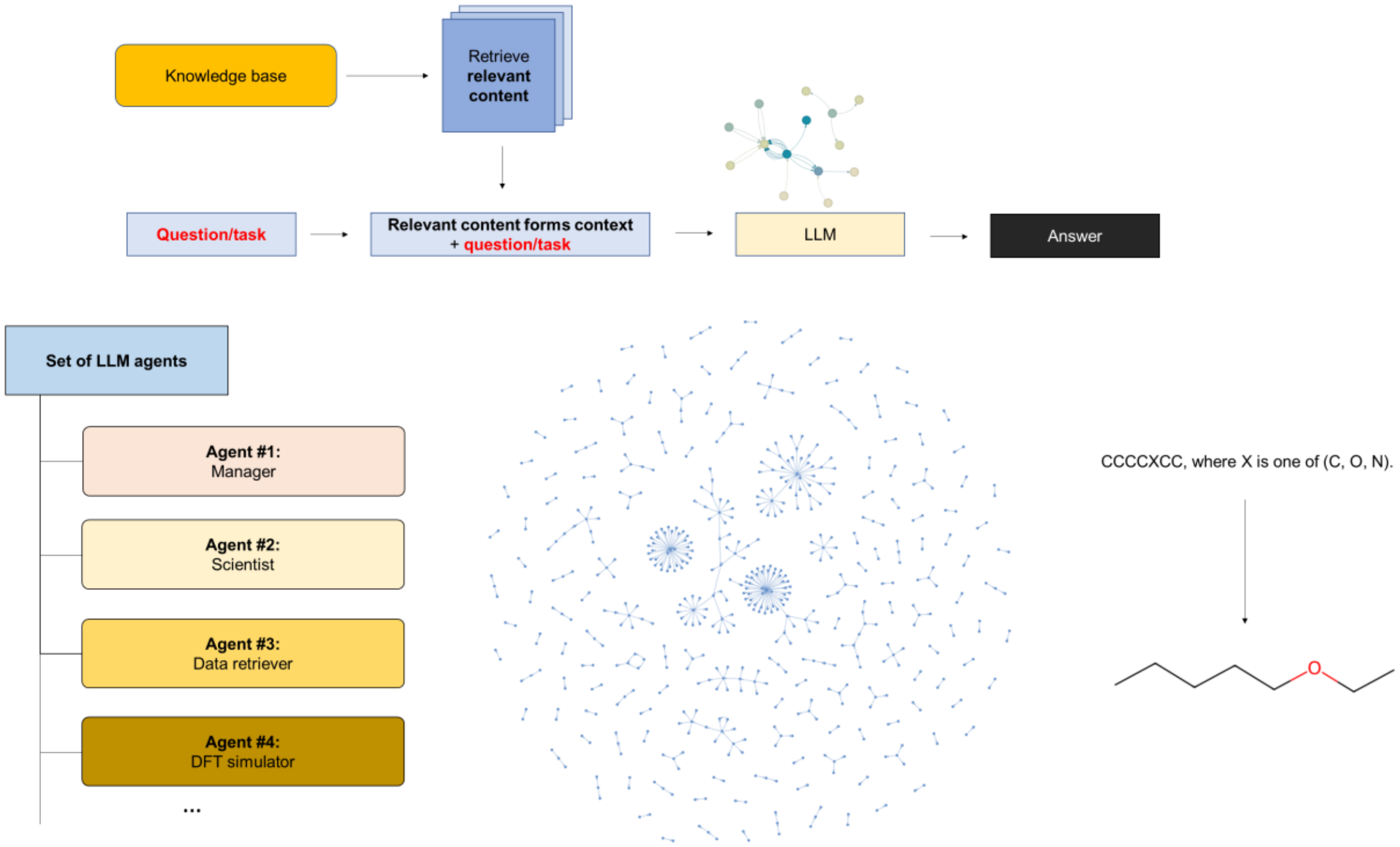
Knowledge base
Retrieve relevant content
Question/task
Relevant content forms context + question/task
LLM
Answer
Set of LLM agents
Agent #1: Manager
Agent #2: Scientist
Agent #3: Data retriever
Agent #4: DFT simulator
...
CCCCXCC, where X is one of (C, O, N).
O